\documentclass{sig-alternate-05-2015}
\usepackage{algorithm}
\usepackage{algorithmic}
\usepackage{makecell}
\usepackage{tablefootnote}
\usepackage{graphicx}
\usepackage{caption}
\usepackage{amsmath,graphicx}
\usepackage{caption}
\usepackage{tablefootnote}
\usepackage{bbm}
\usepackage{xcolor}
\usepackage{amsmath, amssymb}
\usepackage{slashbox}
\usepackage{tablefootnote}
\usepackage{subfigure}
\usepackage{float}
\usepackage{sidecap}
\usepackage{footnote}
\usepackage{multirow}
\usepackage{url}
\usepackage{color}
\usepackage{stfloats}

\begin{document}

% Copyright

\CopyrightYear{2016} 
\setcopyright{acmcopyright}
\conferenceinfo{MM '16,}{October 15-19, 2016, Amsterdam, Netherlands}
\isbn{978-1-4503-3603-1/16/10}\acmPrice{\$15.00}
\doi{http://dx.doi.org/10.1145/2964284.2964299}

%\setcopyright{acmcopyright}
%%\setcopyright{acmlicensed}
%%\setcopyright{rightsretained}
%%\setcopyright{usgov}
%%\setcopyright{usgovmixed}
%%\setcopyright{cagov}
%%\setcopyright{cagovmixed}
%
%
%% DOI
%\doi{10.475/123_4}
%
%% ISBN
%\isbn{123-4567-24-567/08/06}
%
%%Conference
%\conferenceinfo{PLDI '13}{June 16--19, 2013, Seattle, WA, USA}
%
%\acmPrice{\$15.00}

%
% --- Author Metadata here ---
%\conferenceinfo{WOODSTOCK}{'97 El Paso, Texas USA}
%\CopyrightYear{2007} % Allows default copyright year (20XX) to be over-ridden - IF NEED BE.
%\crdata{0-12345-67-8/90/01}  % Allows default copyright data (0-89791-88-6/97/05) to be over-ridden - IF NEED BE.
% --- End of Author Metadata ---

\title{Image Captioning with Deep Bidirectional LSTMs}

\numberofauthors{4} %  in this sample file, there are a *total*
\author{
% 1st. author
%\alignauthor 
Cheng Wang, Haojin Yang, Christian Bartz, Christoph Meinel\\ 
       \affaddr{Hasso Plattner Institute, University of Potsdam}\\
       \affaddr{Prof.-Dr.-Helmert-Str. 2-3, 14482 Potsdam, Germany}\\
       \email{\{cheng.wang, haojin.yang,christoph.meinel\}}@hpi.de \\
       \email{christian.bartz@student.hpi.uni-potsdam.de}
% 2nd. author
%\alignauthor
%G.K.M. Tobin
%       \affaddr{Institute for Clarity in Documentation}\\
%       \affaddr{P.O. Box 1212}\\
%       \affaddr{Dublin, Ohio 43017-6221}\\
%       \email{webmaster@marysville-ohio.com}
%% 3rd. author
%\alignauthor sss
%       \affaddr{The Th{\o}rv{\"a}ld Group}\\
%       \affaddr{1 Th{\o}rv{\"a}ld Circle}\\
%       \affaddr{Hekla, Iceland}\\
%       \email{larst@affiliation.org} 
%\and
%\alignauthor sss
%       \affaddr{The Th{\o}rv{\"a}ld Group}\\
%       \affaddr{1 Th{\o}rv{\"a}ld Circle}\\
%       \affaddr{Hekla, Iceland}\\
%       \email{larst@affiliation.org}% use '\and' if you need 'another row' of author names
}
% There's nothing stopping you putting the seventh, eighth, etc.
% author on the opening page (as the 'third row') but we ask,
% for aesthetic reasons that you place these 'additional authors'
% in the \additional authors block, viz.
%\additionalauthors{Additional authors: John Smith (The Th{\o}rv{\"a}ld Group,
%email: {\texttt{jsmith@affiliation.org}}) and Julius P.~Kumquat
%(The Kumquat Consortium, email: {\texttt{jpkumquat@consortium.net}}).}
% Just remember to make sure that the TOTAL number of authors
% is the number that will appear on the first page PLUS the
% number that will appear in the \additionalauthors section.

\maketitle
\begin{abstract}
This work presents an end-to-end trainable deep bidirectional LSTM (Long-Short Term Memory) model for image captioning. Our model builds on a deep convolutional neural network (CNN) and two separate LSTM networks. It is capable of learning long term visual-language interactions by making use of history and future context information at high level semantic space. Two novel deep bidirectional variant models, in which we increase the depth of nonlinearity transition in different way, are proposed to learn hierarchical visual-language embeddings. Data augmentation techniques such as multi-crop, multi-scale and vertical mirror are proposed to prevent overfitting in training deep models. We visualize the evolution of bidirectional LSTM internal states over time and qualitatively analyze how our models ``translate'' image to sentence. Our proposed models are evaluated on caption generation and image-sentence retrieval tasks with three benchmark datasets: Flickr8K, Flickr30K and MSCOCO datasets. We demonstrate that bidirectional LSTM models achieve highly competitive performance to the state-of-the-art results on caption generation even without integrating additional mechanism (e.g. object detection, attention model etc.) and significantly outperform recent methods on retrieval task.
\end{abstract}



%
%  Use this command to print the description
%

% We no longer use \terms command


\section{Introduction}
Automatically describe an image using sentence-level captions has been receiving much attention recent years \cite{karpathy2015deep,karpathy2014deep,kiros2014unifying,kuznetsova2014treetalk,kuznetsova2012collective,mao2014deep,socher2014grounded,vinyals2015show}. 
It is a challenging task integrating visual and language understanding. 
It requires not only the recognition of visual objects in an image and the semantic interactions between objects, but the ability to capture visual-language interactions and learn how to ``translate'' the visual understanding to sensible sentence descriptions. 
The most important part of this visual-language modeling is to capture the semantic correlations across image and sentence by learning a multimodal joint model. 
While some previous models \cite{li2011composing,kulkarni2013babytalk,mitchell2012midge,kuznetsova2014treetalk,kuznetsova2012collective} have been proposed to address the problem of image captioning, they rely on either use sentence templates, or treat it as retrieval task through ranking the best matching sentence in database as caption. Those approaches usually suffer difficulty in generating variable-length and novel sentences. 
Recent work \cite{karpathy2015deep,karpathy2014deep,kiros2014unifying,mao2014deep,socher2014grounded,vinyals2015show} has indicated that embedding visual and language to common semantic space with relatively shallow recurrent neural network (RNN) can yield promising results. %Unfortunately, those models are limited to capture both long range history and future context in predict sentence word by word over time. 

In this work, we propose novel architectures to the problem of image captioning. Different to previous models, we learn a visual-language space where sentence embeddings are encoded using bidirectional Long-Short Term Memory (Bi-LSTM) and visual embeddings are encoded with CNN. Bi-LSTM is able to summarize long range visual-language interactions from forward and backward directions. Inspired by the architectural depth of human brain, we also explore the deep bidirectional LSTM architectures to learn higher level visual-language embeddings. All proposed models can be trained in end-to-end by optimizing a joint loss.

\begin{figure}[t!]
\centering
  \subfigure{
    \includegraphics[width=0.48\textwidth]{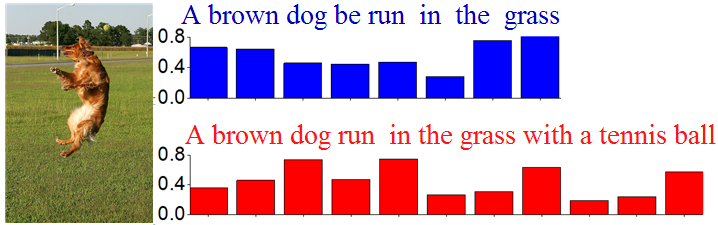}}
  \subfigure{
    \includegraphics[width=0.48\textwidth]{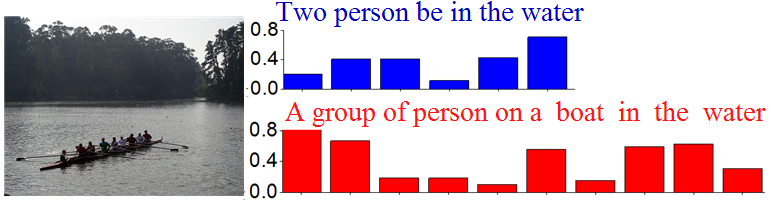}}
\caption{Illustration of generated captions. Two example images from Flickr8K dataset and their best matching captions that generated in forward order (blue) and backward order (red). Bidirectional models capture different levels of visual-language interactions (more evidence see Sec.\ref{sec.vis}). The final caption is the sentence with higher probabilities (histogram under sentence). In both examples, backward caption is selected as final caption for corresponding image.}
\label{illustration_bidirectional_examples}
\end{figure}

\textit{Why bidirectional LSTMs?} In unidirectional sentence generation, one general way of predicting next word $w_t$ with visual context $I$ and history textual context $w_{1:t-1}$ is maximize $\log P(w_t|I,w_{1:t-1})$. While unidirectional model includes past context, it is still limited to retain future context $w_{t+1:T}$ that can be used for reasoning previous word $w_t$ by maximizing $\log P(w_t|I,w_{t+1:T})$. Bidirectional model tries to overcome the shortcomings that each unidirectional (forward and backward direction) model suffers on its own and exploits the past and future dependence to give a prediction. As shown in Figure \ref{illustration_bidirectional_examples}, two example images with bidirectionally generated sentences intuitively support our assumption that bidirectional captions are complementary, combining them can generate more sensible captions.

\textit{Why deeper LSTMs?} The recent success of deep CNN in image classification and object detection \cite{krizhevsky2012imagenet,simonyan2014very} demonstrates that deep, hierarchical models can be more efficient at learning representation than shallower ones. 
This motivated our work to explore deeper LSTM architectures in the context of learning bidirectional visual-language embeddings. As claimed in \cite{pascanu2013construct}, if we consider LSTM as a composition of multiple hidden layers that unfolded in time, LSTM is already deep network. But this is the way of increasing ``horizontal depth'' in which network weights $W$ are reused at each time step and limited to learn more representative features as increasing the ``vertical depth'' of network. To design deep LSTM, one straightforward way is to stack multiple LSTM layers as hidden to hidden transition. 
%It allows to learn higher level representation from input data.
Alternatively, instead of stacking multiple LSTM layers, we propose to add multilayer perception (MLP) as intermediate transition between LSTM layers. This can not only increase LSTM network depth, but can also prevent the parameter size from growing dramatically. 

The core contributions of this work are threefold:
\begin{itemize}
\item We propose an end-to-end trainable multimodal bidirectional LSTM (see Sec.\ref{sec.bi-lstm}) and its deeper variant models (see Sec.\ref{sec.deep-lstm}) that embed image and sentence into a high level semantic space by exploiting both long term history and future context. 
\item We visualize the evolution of hidden states of bidirectional LSTM units to qualitatively analyze and understand how to generate sentence that conditioned by visual context information over time (see Sec.\ref{sec.vis}). 
\item We demonstrate the effectiveness of proposed models on three benchmark datasets: Flickr8K, Flickr30K and MSCOCO. Our experimental results show that bidirectional LSTM models achieve highly competitive performance to the state-of-the-art on caption generation (see Sec.\ref{sec.com_generation}) and perform significantly better than recent methods on retrieval task (see Sec.\ref{sec.retrieval}).
\end{itemize}

\section{Related Work}
Multimodal representation learning \cite{ngiam2011multimodal,srivastava2012multimodal} has significant value in multimedia understanding and retrieval. The shared concept across modalities plays an important role in bridging the ``semantic gap'' of multimodal data. Image captioning falls into this general category of learning multimodal representations. 

Recently, several approaches have been proposed for image captioning. We can roughly classify those methods into three categories. The first category is template based approaches that generate caption templates based on detecting objects and discovering attributes within image. For example, the work \cite{li2011composing} was proposed to parse a whole sentence into several phrases, and learn the relationships between phrases and objects within an image. In \cite{kulkarni2013babytalk}, conditional random field (CRF) was used to correspond objects, attributes and prepositions of image content and predict the best label.  Other similar methods were presented in \cite{mitchell2012midge,kuznetsova2014treetalk,kuznetsova2012collective}. These methods are typically hard-designed and rely on fixed template, which mostly lead to poor performance in generating variable-length sentences. The second category is retrieval based approach, this sort of methods treat image captioning as retrieval task. By leveraging distance metric to retrieve similar captioned images, then modify and combine retrieved captions to generate caption \cite{kuznetsova2014treetalk}. But these approaches generally need additional procedures such as modification and generalization process to fit image query.

Inspired by the success use of CNN \cite{krizhevsky2012imagenet,zeiler2014visualizing} and Recurrent Neural Network \cite{mikolov2010recurrent,mikolov2011extensions,bahdanau2014neural}. The third category is emerged as neural network based methods \cite{vinyals2015show,xu2015show,kiros2014unifying,karpathy2014deep,karpathy2015deep}. Our work also belongs to this category. Among those work, Kiro \emph{et al.}\cite{kiros2014multimodal} can been as pioneer work to use neural network for image captioning with multimodal neural language model. In their follow up work \cite{kiros2014unifying}, Kiro \emph{et al.} introduced an encoder-decoder pipeline where sentence was encoded by LSTM and decoded with structure-content neural language model (SC-NLM).  Socher \emph{et al}.\cite{socher2014grounded} presented a DT-RNN (Dependency Tree-Recursive Neural Network) to embed sentence into a vector space in order to retrieve images. Later on, Mao \emph{et al}.\cite{mao2014deep} proposed m-RNN which replaces feed-forward neural language model in \cite{kiros2014unifying}. Similar architectures were introduced in NIC \cite{vinyals2015show} and LRCN \cite{donahue2015long}, both approaches use LSTM to learn text context. But NIC only feed visual information at first time step while Mao \emph{et al}.\cite{mao2014deep} and LRCN \cite{donahue2015long}'s model consider image context at each time step. Another group of neural network based approaches has been introduced in \cite{karpathy2014deep,karpathy2015deep} where image captions generated by integrating object detection with R-CNN (region-CNN) and inferring the alignment between image regions and descriptions.  

Most recently, Fang \emph{et al}.\cite{fang2015captions} used multi-instance learning and traditional maximum-entropy language model for description generation. Chen \emph{et al}.\cite{chen2015mind} proposed to learn visual representation with RNN for generating image caption. In \cite{xu2015show}, Xu \emph{et al}. introduced attention mechanism of human visual system into encoder-decoder framework. It is shown that attention model can visualize what the model ``see'' and yields significant improvements on image caption generation. Unlike those models, our deep LSTM model directly assumes the mapping relationship between visual-language is antisymmetric and dynamically learns long term bidirectional and hierarchical visual-language interactions. This is proved to be very effective in generation and retrieval tasks as we demonstrate in Sec.\ref{sec.com_generation} and Sec.\ref{sec.retrieval}.
\section{Model}
In this section, we describe our multimodal bidirectional LSTM model (Bi-LSTM for short) and explore its deeper variants. We first briefly introduce LSTM which is at the center of model. The LSTM we used is described in \cite{zaremba2014learning}.
\subsection{Long Short Term Memory}
\begin{figure}
\centering
\includegraphics[width=0.22\textwidth]{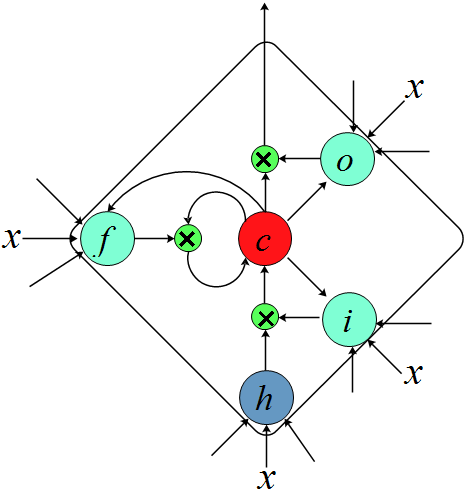}
\caption{\label{fig:lstm} Long Short Term Memory (LSTM) cell. It is consist of an input gate $i$, a forget gate $f$, a memory cell $c$ and an output gate $o$. The input gate decides let incoming signal go through to memory cell or block it. The output gate can allow new output or prevent it. The forget gate decides to remember or forget cell's previous state. Updating cell states is performed by feeding previous cell output to itself by recurrent connections in two consecutive time steps.}
\end{figure}

Our model builds on LSTM cell, which is a particular form of traditional recurrent neural network (RNN). It has been successfully applied to machine translation \cite{cho2014learning}, speech recognition \cite{graves2013speech} and sequence learning \cite{sutskever2014sequence}. As shown in Figure \ref{fig:lstm}, the reading and writing memory cell $c$ is controlled by
a group of sigmoid gates. At given time step $t$, LSTM receives inputs from different sources: current input $\mathbf{x}_t$, the previous hidden state of all LSTM units $\mathbf{h}_{t-1}$ as well as previous memory cell state $\mathbf{c}_{t-1}$. The updating of those gates at time step $t$ for given inputs $\mathbf{x}_t$, $\mathbf{h}_{t-1}$ and $\mathbf{c}_{t-1}$ as follows.
\begin{align}
\mathbf{i}_t=\sigma (\mathbf{W}_{xi}\mathbf{x}_t+\mathbf{W}_{hi}\mathbf{h}_{t-1}+\mathbf{b}_i)\\
\mathbf{f}_t=\sigma (\mathbf{W}_{xf}\mathbf{x}_t+\mathbf{W}_{hf}\mathbf{h}_{t-1}+\mathbf{b}_f)\\
\mathbf{o}_t=\sigma (\mathbf{W}_{xo}\mathbf{x}_t+\mathbf{W}_{ho}\mathbf{h}_{t-1}+\mathbf{b}_o)\\
\mathbf{g}_t=\phi (\mathbf{W}_{xc}\mathbf{x}_t+\mathbf{W}_{hc}\mathbf{h}_{t-1}+\mathbf{b}_c)\\
\mathbf{c}_t=\mathbf{f}_t\odot\mathbf{c}_{t-1}+\mathbf{i}_t\odot\mathbf{g}_t\\
\mathbf{h}_t=\mathbf{o}_t\odot\phi(\mathbf{c}_t)
\end{align}
where $\mathbf{W}$ are the weight matrices learned from the network and $\mathbf{b}$ are bias vectors. $\sigma$ is the sigmoid activation function $\sigma(x)=1/(1+\exp(-x))$ and $\phi$ presents hyperbolic tangent $\phi(x)=(\exp(x)-\exp(-x))/(\exp(x)+\exp(-x))$. 
$\odot$ denotes the products with a gate value. 
The LSTM hidden output $\mathbf{h}_t$=$\{\mathbf{h}_{tk}\}_{k=0}^K$, $\mathbf{h}_t\in \mathbf{R}^K$ will be used to predict the next word by Softmax function with parameters $\mathbf{W}_s$ and $\mathbf{b}_s$:
\begin{equation}
\mathcal{F}(\mathbf{p}_{ti};\mathbf{W}_s,\mathbf{b}_s)=\frac{\exp(\mathbf{W}_s\mathbf{h}_{ti}+\mathbf{b}_s)}{\sum_{j=1}^{K}\exp(\mathbf{W}_s\mathbf{h}_{tj}+\mathbf{b}_s)}
\end{equation}
where $\mathbf{p}_{ti}$ is the probability distribution for predicted word.% and $\mathbf{W}_s$ and $b$ are weights and bias term of Softmax layer.

Our key motivation of chosen LSTM is that it can learn long-term temporal activities and avoid quick exploding and vanishing problems that traditional RNN suffers from during back propagation optimization.

\begin{figure}
\centering
\includegraphics[width=0.46\textwidth]{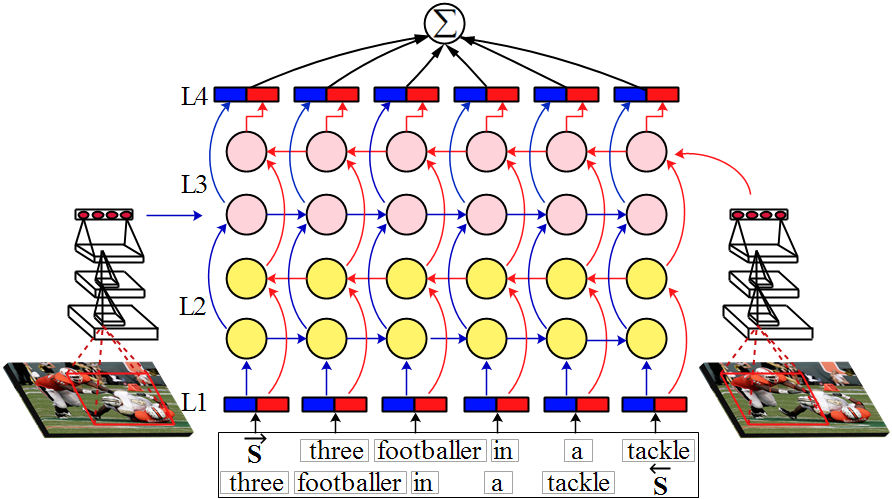}
\caption{\label{fig:framework} Multimodal Bidirectional LSTM. L1: sentence embedding layer. L2: T-LSTM layer. L3: M-LSTM layer. L4: Softmax layer. We feed sentence in both forward (blue arrows) and backward (red arrows) order which allows our model summarizes context information from both left and right side for generating sentence word by word over time. Our model is end-to-end trainable by minimize a joint loss.}\;
\end{figure}

\begin{figure*}
\centering
\includegraphics[width=1\textwidth]{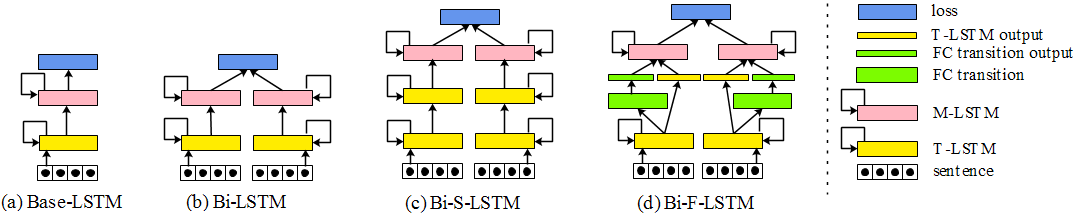}
\caption{Illustrations of proposed deep architectures for image captioning. The network in (a) is commonly used in previous work, e.g. \cite{donahue2015long,mao2014deep}. (b) Our proposed Bidirectional LSTM (Bi-LSTM). (c) Our proposed Bidirectional Stacked LSTM (Bi-S-LSTM). (d) Our proposed Bidirectional LSTM with fully connected (FC) transition layer (Bi-F-LSTM).}
\label{fig:extension_LSTM1}
\end{figure*}

\subsection{Bidirectional LSTM}
\label{sec.bi-lstm}
In order to make use of both the past and future context information of a sentence in predicting word, we propose a bidirectional model by feeding sentence to LSTM from forward and backward order. 
Figure \ref{fig:framework} presents the overview of our model, it is comprised of three modules: a CNN for encoding image inputs, a Text-LSTM (T-LSTM) for encoding sentence inputs, a Multimodal LSTM (M-LSTM) for embedding visual and textual vectors to a common semantic space and decoding to sentence. The bidirectional LSTM is implemented with two separate LSTM layers for computing forward hidden sequences $\overrightarrow{\mathbf{h}}$ and backward hidden sequences $\overleftarrow{\mathbf{h}}$. The forward LSTM starts
at time $t=1$ and the backward LSTM starts at time $t=T$. Formally, our model works as follows, for raw image input $\widetilde{I}$, forward order sentence$\overrightarrow{S}$ and backward order sentence  $\overleftarrow{S}$, the encoding performs as
\begin{equation}
\begin{aligned}
\mathbf{I}_t=\mathcal{C}(\widetilde{I};\mathbf{\Theta}_v)~~
\overrightarrow{\mathbf{h}}_t^1=\mathcal{T}(\overrightarrow{\mathbf{E}}\overrightarrow{\mathbf{S}};\overrightarrow{\mathbf{\Theta}}_l)~~ \overleftarrow{\mathbf{h}}_t^1=\mathcal{T}(\overleftarrow{\mathbf{E}}\overleftarrow{\mathbf{S}};\overleftarrow{\mathbf{\Theta}}_l)
\end{aligned}
\end{equation}
where $\mathcal{C}$, $\mathcal{T}$ represent CNN, T-LSTM respectively and $\mathbf{\Theta}_v$, $\mathbf{\Theta}_l$ are their corresponding weights. $\overrightarrow{\mathbf{E}}$ and $\overleftarrow{\mathbf{E}}$ are bidirectional embedding matrices learned from network. 
Encoded visual and textual representations are then embedded to multimodal LSTM by: 
\begin{equation}
\begin{aligned}
\overrightarrow{\mathbf{h}}_t^2=\mathcal{M}(\overrightarrow{\mathbf{h}}_t^1,\mathbf{I}_t;\overrightarrow{\mathbf{\Theta}}_m)~~~~ \overleftarrow{\mathbf{h}}_t^2=\mathcal{M}(\overleftarrow{\mathbf{h}}_t^1,\mathbf{I}_t;\overleftarrow{\mathbf{\Theta}}_m)\\
\end{aligned}
\end{equation}
where $\mathcal{M}$ presents M-LSTM and its weight $\mathbf{\Theta}_m$.  $\mathcal{M}$ aims to capture the correlation of visual context and words at different time steps. We feed visual vector $\mathbf{I}$ to model at each time step for capturing strong visual-word correlation. On the top of M-LSTM are Softmax layers which compute the probability distribution of next predicted word by
\begin{equation}
\begin{aligned}
\overrightarrow{\mathbf{p}}_{t+1}=\mathcal{F}(\overrightarrow{\mathbf{h}}_t^2;\mathbf{W}_s,\mathbf{b}_s)~~~~ \overleftarrow{\mathbf{p}}_{t+1}=\mathcal{F}(\overleftarrow{\mathbf{h}}_t^2;\mathbf{W}_s,\mathbf{b}_s)\\
\end{aligned}
\end{equation}
where $\mathbf{p}\in \mathbf{R}^K$ and $K$ is the vocabulary size.
 
\subsection{Deeper LSTM architecture}
\label{sec.deep-lstm}
\begin{figure}
\centering
\includegraphics[width=0.35\textwidth]{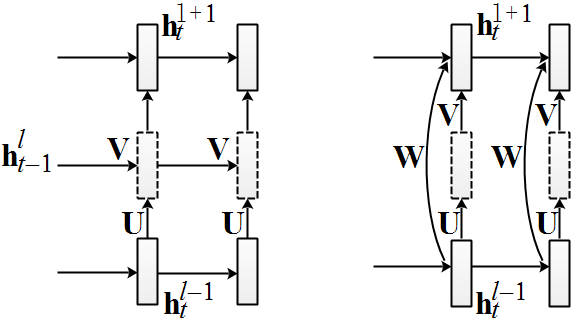}
\caption{Transition for Bi-S-LSTM(L) and Bi-F-LSTM(R)}
\label{fig:transition_function}
\end{figure}

To design deeper LSTM architectures, in addition to directly stack multiple LSTMs on each other that we named as Bi-S-LSTM (Figure \ref{fig:extension_LSTM1}(c)), we propose to use a fully connected layer as intermediate transition layer. Our motivation comes from the finding of \cite{pascanu2013construct}, in which DT(S)-RNN (deep transition RNN with shortcut) is designed by adding hidden to hidden multilayer perception (MLP) transition. It is arguably easier to train such network. Inspired by this, we extend Bi-LSTM (Figure \ref{fig:extension_LSTM1}(b)) with a fully connected layer that we called Bi-F-LSTM (Figure \ref{fig:extension_LSTM1}(d)), shortcut connection between input and hidden states is introduced to make it easier to train model. The aim of extension models is to learn an extra hidden transition function $F_h$. In Bi-S-LSTM
\begin{equation}
\mathbf{h}_t^{l+1}=F_h(\mathbf{h}_t^{l-1},\mathbf{h}_{t-1}^l)=\mathbf{U}\mathbf{h}_t^{l-1}+\mathbf{V}\mathbf{h}_{t-1}^l
\end{equation}
where $\mathbf{h}_t^l$ presents the hidden states of $l$-th layer at time $t$, $\mathbf{U}$ and $\mathbf{V}$ are matrices connect to transition layer (also see Figure \ref{fig:transition_function}(L)).  For readability, we consider one direction training and suppress bias terms. 
Similarly, in Bi-F-LSTM, to learn a hidden transition function $F_h$ by
\begin{equation}
\mathbf{h}_t^{l+1}=F_h(\mathbf{h}_t^{l-1})=\phi_r(\mathbf{W}\mathbf{h}_t^{l-1}\oplus(\mathbf{V}(\mathbf{U}\mathbf{h}_t^{l-1}))
\end{equation}
where $\oplus$ is the operator that concatenates $\mathbf{h}_t^{l-1}$ and its abstractions to a long hidden states (also see Figure \ref{fig:transition_function}(R)). $\phi_r$ presents rectified linear unit (Relu) activation function for transition layer, which performs $\phi_r(x)=\max(0,x)$.

\subsection{Data Augmentation}
One of the most challenging aspects of training deep bidirectional LSTM models is preventing overfitting. Since our largest dataset has only 80K images \cite{lin2014microsoft} which might cause overfitting easily, we adopted several techniques such as fine-tuning on pre-trained visual model, weight decay, dropout and early stopping that commonly used in the literature. Additionally, it has been proved that data augmentation such as randomly cropping and horizontal mirror \cite{simonyan2014two,lu2014rapid}, adding noise, blur and rotation \cite{Wang:2015:DIY:2733373.2806219} can effectively alleviate over-fitting and other. Inspired by this, we designed new data augmentation techniques to increase the number of image-sentence pairs. 
Our implementation performs on visual model, as follows:  
\begin{itemize}
\item \textbf{Multi-Corp}: Instead of randomly cropping on input image, we crop at the four corners and center region. Because we found random cropping is more tend to select center region and cause overfitting easily. By cropping four corners and center, the variations of network input can be increased to alleviate overfitting.
\item \textbf{Multi-Scale}: To further increase the number of image-sentence pairs, we rescale input image to multiple scales. For each input image $\widetilde{I}$ with size $H\times W$, it is resized to 256 $\times$ 256, then we randomly select a region with size of $s*H\times s*W$, where $s\in[1,0.925,0.875,0.85]$ is scale ratio. $s=1$ means we do not multi-scale operation on given image. Finally we resize it to AlexNet input size 227 $\times$227 or VggNet input size 224 $\times$ 224. 
\item \textbf{Vertical Mirror}: Motivated by the effectiveness of widely used horizontal mirror, it is natural to also consider the vertical mirror of image for same purpose. 
\end{itemize}

Those augmentation techniques are implemented in real-time fashion. Each input image is randomly transformed using one of augmentations to network input for training. In principle, our data augmentation can increase image-sentence training pairs by roughly 40 times (5$\times$4$\times$2).

\subsection{Training and Inference}
Our model is end-to-end trainable by using Stochastic Gradient Descent (SGD). The joint loss function $L=\overrightarrow{L}+\overleftarrow{L}$ is computed by accumulating the Softmax losses of forward and backward directions. Our objective is to minimize $L$, which is equivalent to maximize the probabilities of correctly generated sentences. We compute the gradient $\triangledown L$ with Back-Propagation Through Time (BPTT) algorithm.

The trained model is used to predict a word $w_t$ with given image context $I$ and previous word context $w_{1:{t-1}}$ by $P(w_t|w_{1:{t-1}},I)$ in forward order, or by $P(w_t|w_{{t+1:T}},I)$ in backward order. We set $w_1$=$w_T$=$\mathbf{0}$ at start point respectively for forward and backward directions. Ultimately, with generated sentences from two directions, we decide the final sentence for given image $p(w_{1:T}|I)$ according to the summation of word probability within sentence
\begin{align}
\label{equ:final_caption}
&p(w_{1:T}|I)=\max({\sum\nolimits_{t=1}^{T}}(\overrightarrow{p}(w_{t}|I)), \sum\nolimits_{t=1}^T(\overleftarrow{p}(w_{t}|I))) \\
&\overrightarrow{p}(w_{t}|I)=\prod\nolimits_{t=1}^Tp(w_t|w_{1},w_{2},...,w_{t-1},I) \\
&\overleftarrow{p}(w_{t}|I)=\prod\nolimits_{t=1}^Tp(w_t|w_{t+1},w_{t+2},...,w_{T},I) 
\end{align}

Follow previous work, we adopted beam search to consider the best $k$ candidate sentences at time $t$ to infer the sentence at next time step. In our work, we fix $k=1$ in all experiments although the average of 2 BLEU \cite{papineni2002bleu} points better results can be achieved with $k=20$ compare to $k=1$ as reported in \cite{vinyals2015show}. 

\section{Experiments}
In this section, we design several groups of experiments to accomplish following objectives:
\begin{itemize}
\item Qualitatively analyze and understand how bidirectional multimodal LSTM learns to generate sentence conditioned by visual context information over time.

\item Measure the benefits and performance of proposed bidirectional model and its deeper variant models that we increase their nonlinearity depth from different ways. 

\item Compare our approach with state-of-the-art methods in terms of sentence generation and image-sentence retrieval tasks on popular benchmark datasets. 
\end{itemize}

%%% Dataset%%%
\subsection{Datasets}
To validate the effectiveness, generality and robustness of our models, we conduct experiments on three benchmark datasets: Flickr8K \cite{rashtchian2010collecting}, Flickr30K \cite{young2014image} and MSCOCO \cite{lin2014microsoft}.

\textbf{Flickr8K}. It consists of 8,000 images and each of them has 5 sentence-level captions. We follow the standard dataset divisions provided by authors, 6,000/1,000/1,000 images for training/validation/testing respectively. 

\textbf{Flickr30K}. An extension version of Flickr8K. It has 31,783 images and each of them has 5 captions. We follow the public accessible\footnote{\url {http://cs.stanford.edu/people/karpathy/deepimagesent/}} dataset divisions by Karpathy \emph{et al.} \cite{karpathy2015deep}. In this dataset splits, 29,000/1,000/1,000 images are used for training/validation/testing respectively.

\textbf{MSCOCO}. This is a recent released dataset that covers 82,783 images for training and 40,504 images for validation. Each of images has 5 sentence annotations. Since there is lack of standard splits, we also follow the splits provided by Karpathy \emph{et al}. \cite{karpathy2015deep}. Namely, 80,000 training images and 5,000 images for both validation and testing.
\subsection{Implementation Details}
 \textbf{Visual feature}. We use two visual models for encoding image: Caffe \cite{jia2014caffe} reference model which is pre-trained with AlexNet \cite{krizhevsky2012imagenet} and 16-layer VggNet model \cite{simonyan2014very}. We extract features from last fully connected layer and feed to train visual-language model with LSTM. Previous work \cite{vinyals2015show,mao2014deep} have demonstrated that with more powerful image models such as GoogleNet \cite{szegedy2015going} and VggNet \cite{simonyan2014very} can achieve promising improvements. To make a fair comparison with recent works, we select the widely used two models for experiments.

\textbf{Textual feature}. We first represent each word $w$ within sentence as one-hot vector, $w\in \mathbf{R}^K$, $K$ is vocabulary size built on training sentences and different for different datasets. By performing basic tokenization and removing the words that occurs less than 5 times in the training set, we have 2028, 7400 and 8801 words for Flickr8K, Flickr30K and MSCOCO dataset vocabularies respectively. 

\begin{figure*}[!htb]
\centering
  \subfigure[\scriptsize input]{
    \includegraphics[width=2.5cm,height=5cm]{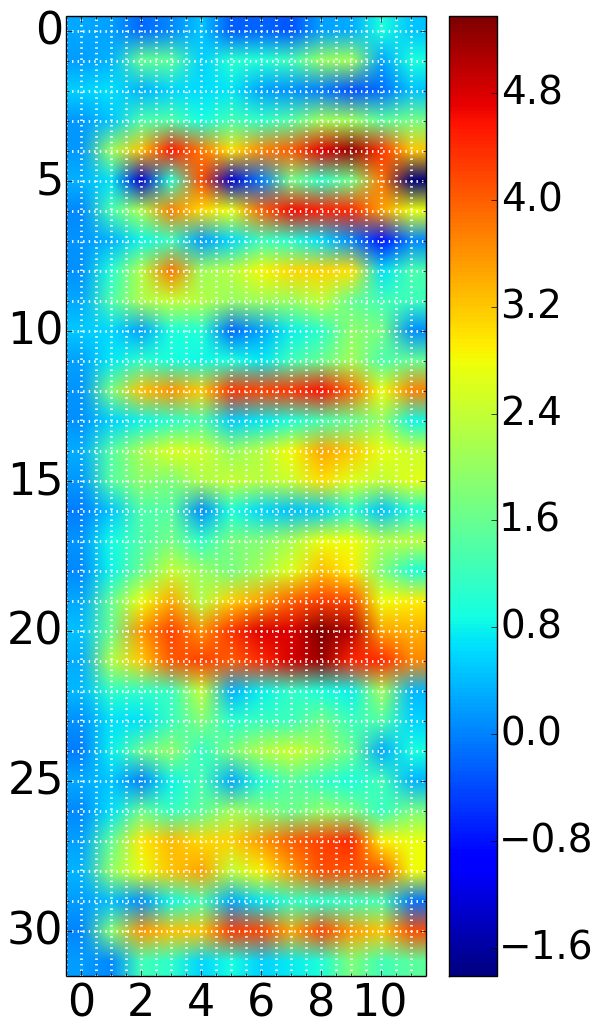}}
  \subfigure[\scriptsize input gate]{
    \includegraphics[width=2.5cm,height=5cm]{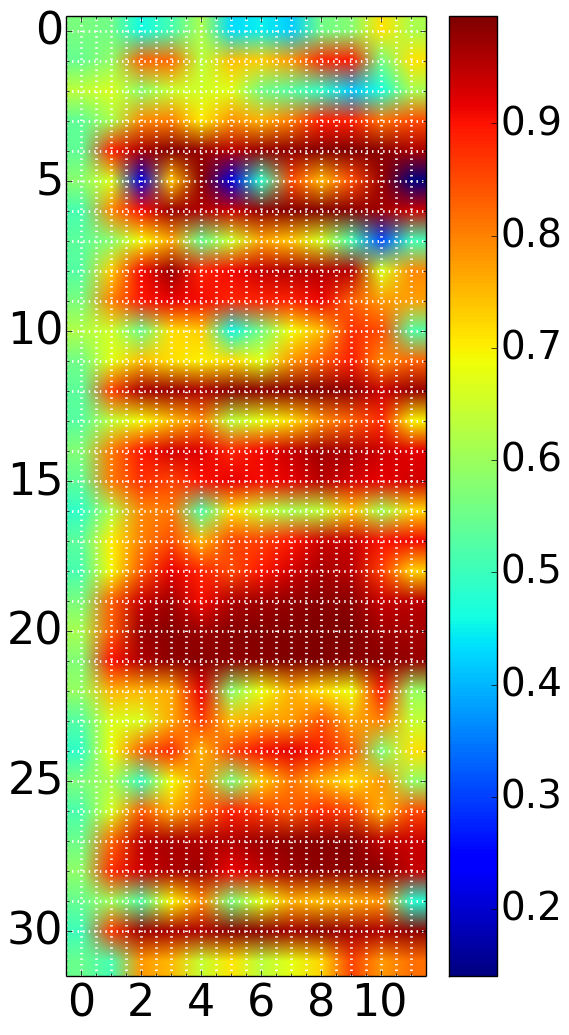}}
\subfigure[\scriptsize forget gate]{
    \includegraphics[width=2.5cm,height=5cm]{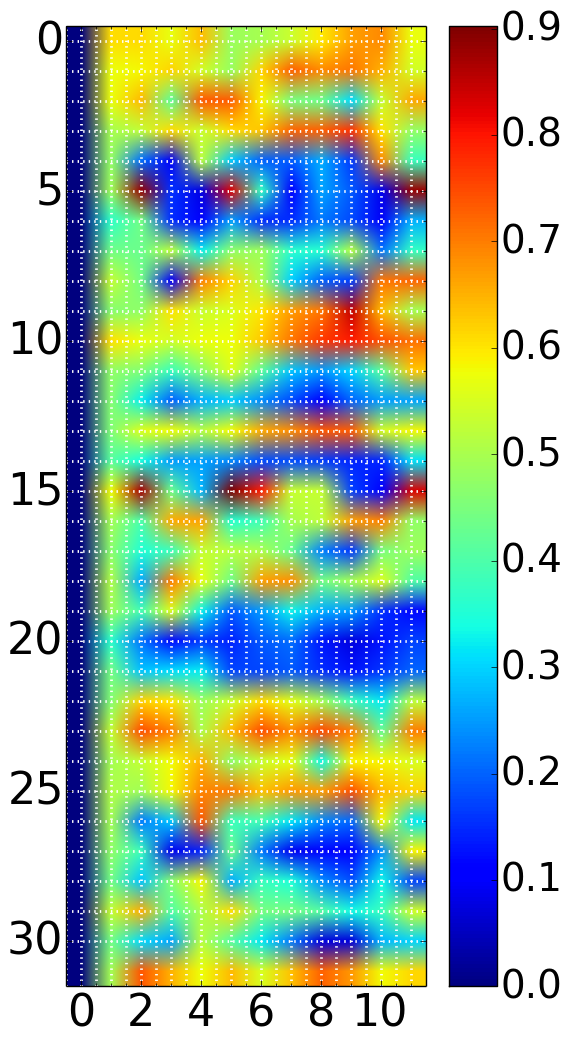}}
\subfigure[\scriptsize cell state]{
    \includegraphics[width=2.5cm,height=5cm]{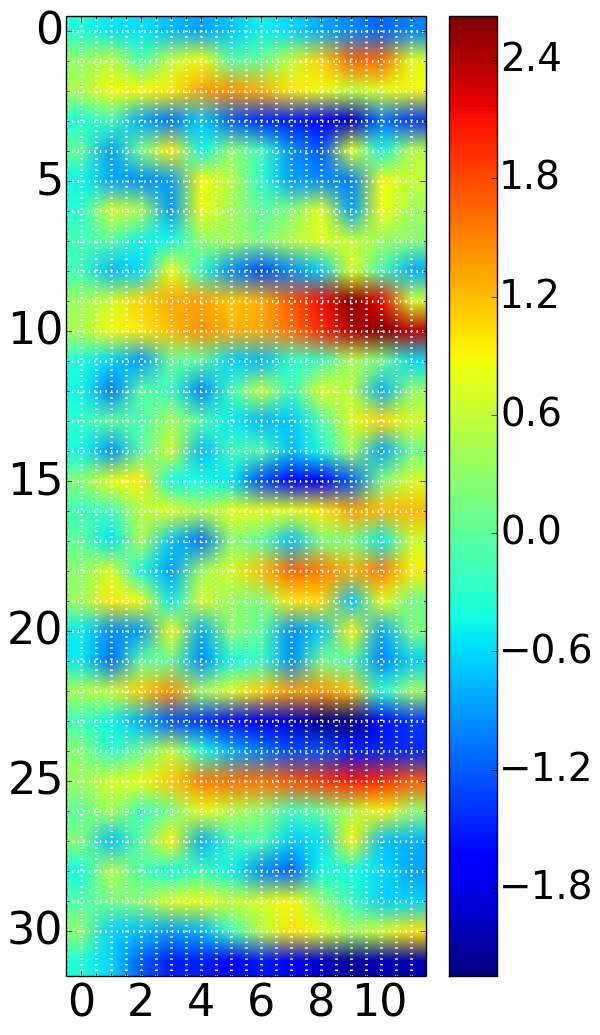}}
\subfigure[\scriptsize output gate]{
    \includegraphics[width=2.5cm,height=5cm]{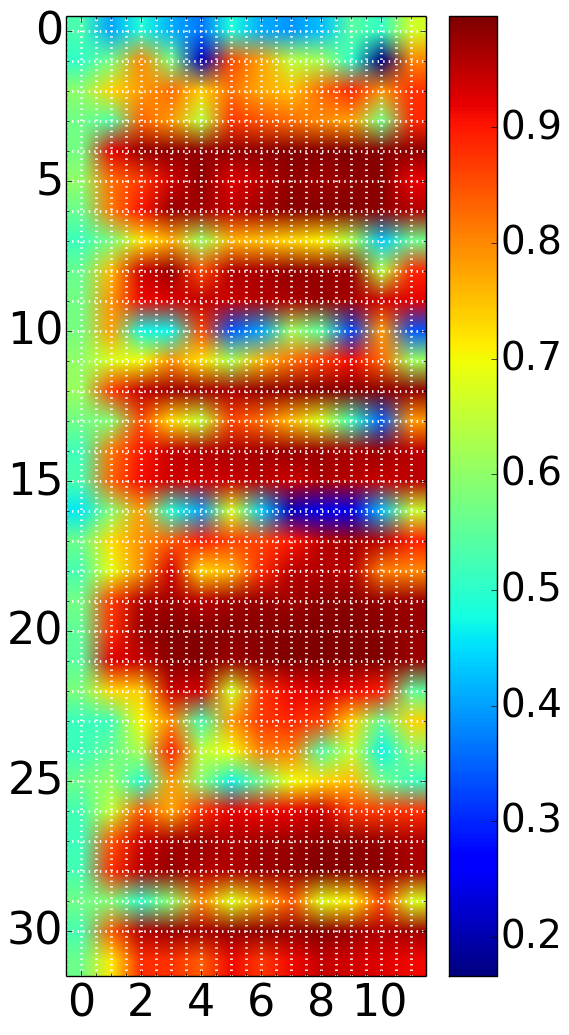}}
\subfigure[\scriptsize output]{
    \includegraphics[width=2.5cm,height=5cm]{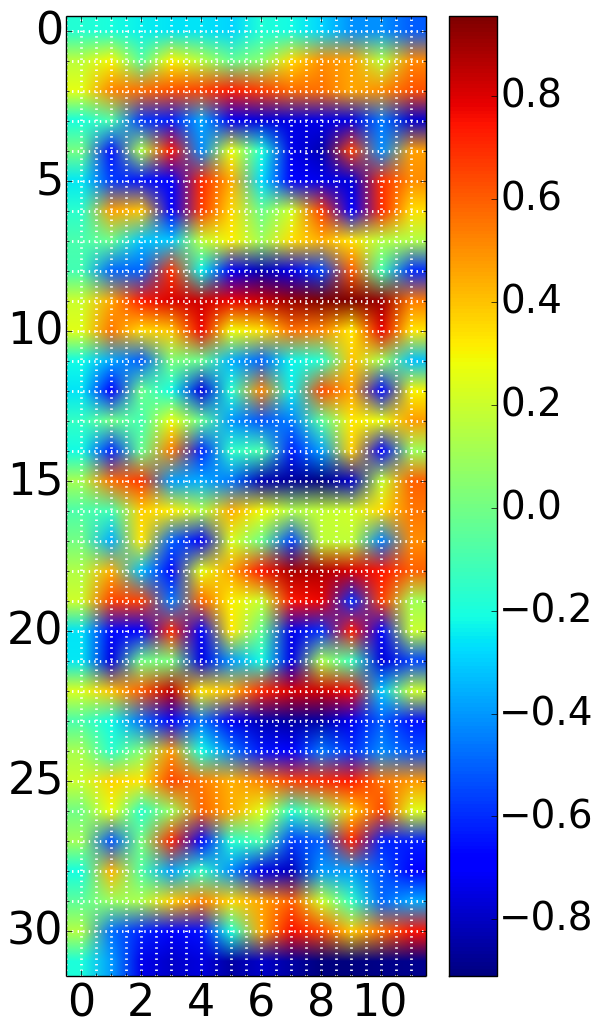}}
\caption{Visualization of LSTM cell. The horizontal axis corresponds to time steps. The vertical axis is cell index. Here we visualize the gates and cell states of the first 32 Bi-LSTM units of T-LSTM in forward direction over 11 time steps.}
\label{fig:gates}
\end{figure*}

\begin{figure*}[!ht]
\centering
  \subfigure[T-LSTM (forward) units]{
    \includegraphics[width=0.48\textwidth]{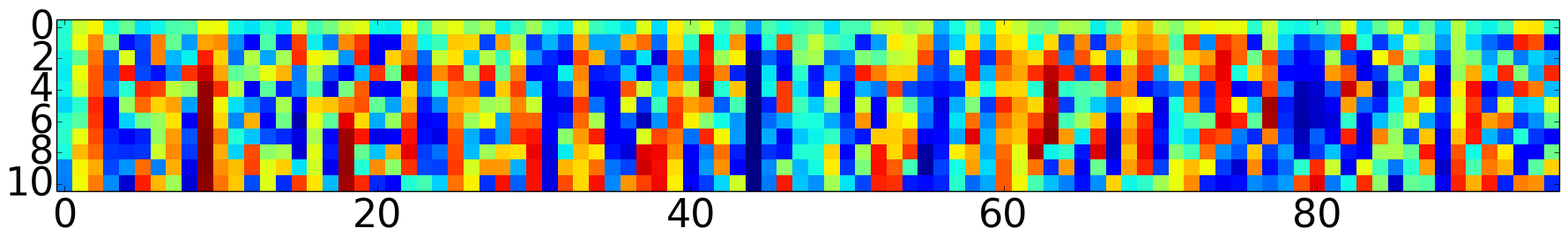}}
  \subfigure[T-LSTM (backward) units]{
    \includegraphics[width=0.48\textwidth]{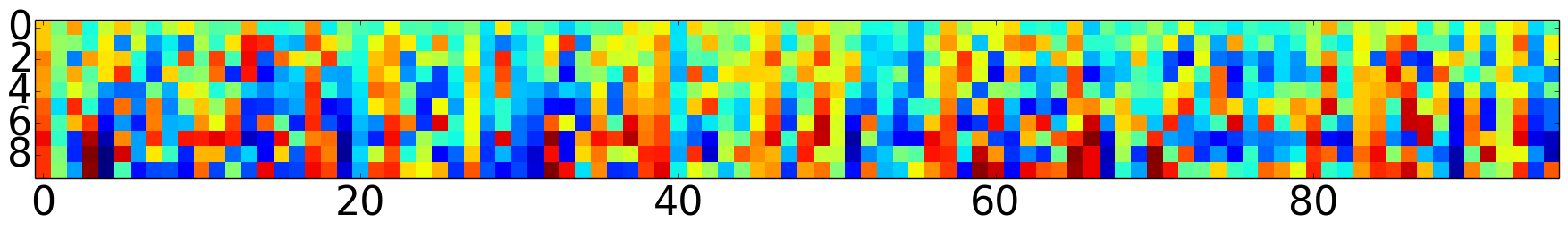}}
\subfigure[M-LSTM (forward) units]{
    \includegraphics[width=0.48\textwidth]{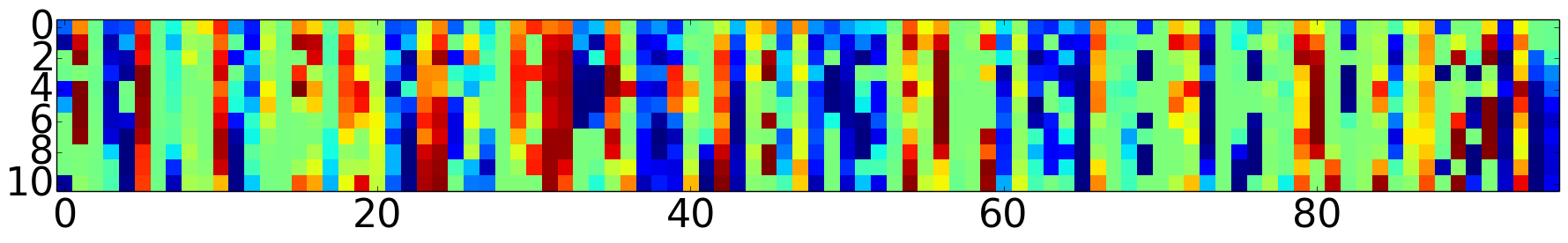}}
\subfigure[M-LSTM (backward) units]{
    \includegraphics[width=0.48\textwidth]{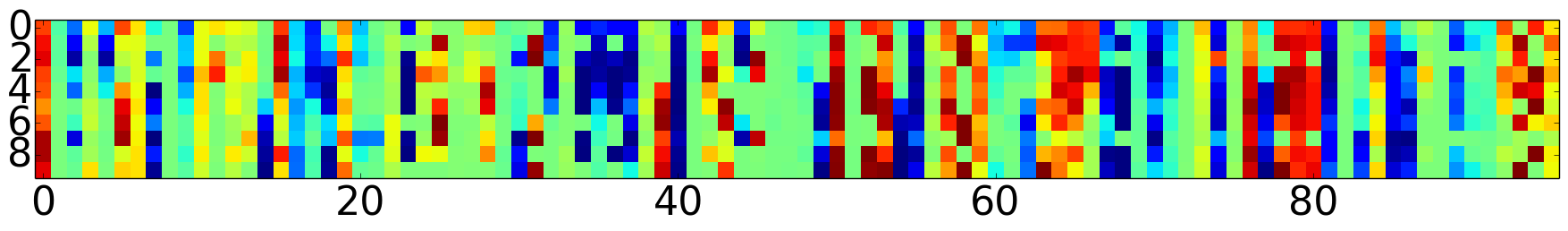}}
\subfigure[probability (forward) units]{
    \includegraphics[width=0.48\textwidth]{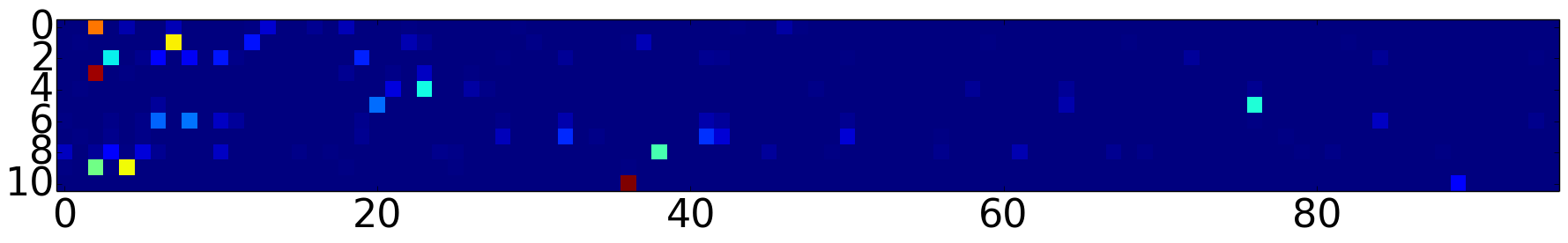}}
\subfigure[probability (backward) units]{
    \includegraphics[width=0.48\textwidth]{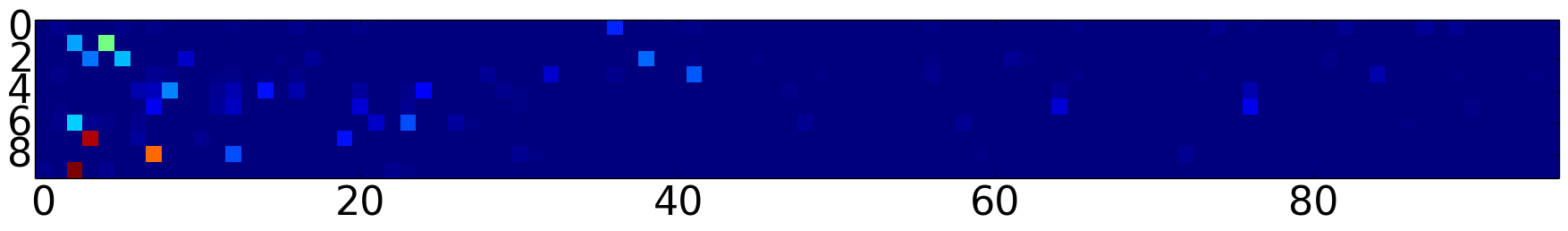}}
\vspace{-1ex}
\centering
\begin{minipage}{1\linewidth} 
\begin{table}[H]
\begin{tabular}{cc}
%\noindent\hfil\rule{0.5\textwidth}{.4pt}
\hline
~A man in a black jacket is walking down the street~~~~~~~~~~~~~~~~~~~ & Street the on walking is suit a in man a~~~~\\~~
2~~~~7  ~~3 ~2 ~~23~~ ~ 76~~~8~~~~~~41~~~~~38~~ ~~4~~~ 36~~~~~~~~~~~~~~~ ~~~~~~ & ~~~36 ~~~4 ~~5~~~~~41~~~~~8~193~~2~3~~~7~~~~2~~~~ \\ 
\hline
\end{tabular}
\centering
\text{~~~~~~~~~~~~~~~~~~~~~~(g) Generated words and corresponding word index in vocabulary}
\end{table}
\end{minipage}
%\vspace{-2ex}
\caption{Pattern of the first 96 hidden units chosen at each layer of Bi-LSTM in both forward and backward directions. The vertical axis presents time steps. The horizontal axis corresponds to different LSTM units. In this example, we visualize the T-LSTM layer for text only, the M-LSTM layer for both text and image and Softmax layer for computing word probability distribution. The model was trained on Flickr 30K dataset for generating sentence word by word at each time step. In (g), we provide the predicted words at different time steps and their corresponding index in vocabulary where we can also read from (e) and (f) (the highlight point at each row). Word with highest probability is selected as the predicted word.}
\label{fig:T-M-LSTM}
\end{figure*}

\begin{figure*}[!htb]
\begin{table}[H]
\begin{tabular}{cccc}
\begin{minipage}{0.22\textwidth}
\centering
\fcolorbox{red}{white!20}{\includegraphics[width=2cm,height=2.5cm]{mscoco_example_1.jpg}}\\ \text{(a)}
\end{minipage} 
&
\begin{minipage}{0.23\textwidth}
\centering
\fcolorbox{red}{white!20}{\includegraphics[width=2.8cm,height=2.5cm]{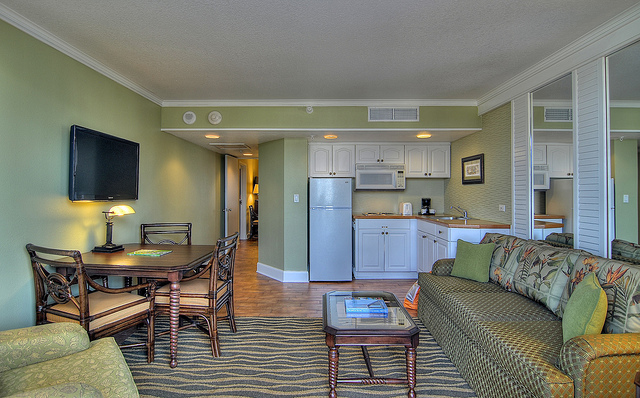}}\\ \text{(b)}
\end{minipage} 
&
\begin{minipage}{0.23\textwidth}
\centering
\fcolorbox{red}{white!20}{\includegraphics[width=2.8cm,height=2.5cm]{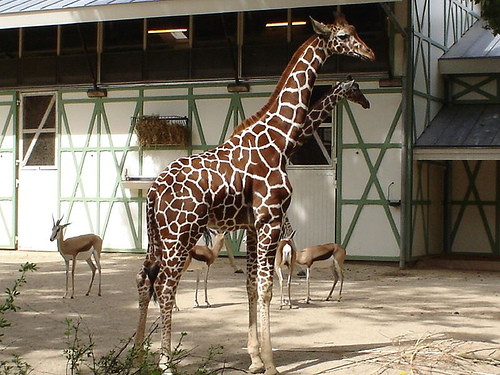}}\\ \text{(c)}
\end{minipage} 
&
\begin{minipage}{0.23\textwidth}
\centering
\fcolorbox{red}{white!20}{\includegraphics[width=2.8cm,height=2.5cm]{mscoco_example_4.jpg}}\\ \text{(d)}
\end{minipage} \\ 

\begin{minipage}{0.22\textwidth}
~\newline
\scriptsize{\textbf{\color{blue}{A woman in a tennis court holding a tennis racket.}}} 
~\newline
\scriptsize{\color{red}{A woman getting ready to hit a tennis ball.}}
\end{minipage}
&
\begin{minipage}{0.23\textwidth}
~\newline
\scriptsize{\textbf{\color{blue}{A living room with a couch and a table.}}} 
~\newline
\scriptsize{\color{red}{Two chairs and a table in a living room.}}
\end{minipage}
&
\begin{minipage}{0.23\textwidth}
~\newline
\scriptsize{\color{blue}{A giraffe standing in a zoo enclosure with a baby in the background.}} 
~\newline
\scriptsize{\textbf{\color{red}{A couple of giraffes are standing at a zoo.}}}
\end{minipage}
&
\begin{minipage}{0.23\textwidth}
~\newline
\scriptsize{\color{blue}{A train is pulling into a train station.}} 
~\newline
\scriptsize{\textbf{\color{red}{A train on the tracks at a train station.}}}
\end{minipage}
\end{tabular}
\end{table}
\vspace{-3ex}
\caption{Examples of generated captions for given query image on MSCOCO validation set. Blue-colored captions are generated in forward direction and red-colored captions are generated in backward direction. The final caption is selected according to equation (\ref{equ:final_caption}) which selects the sentence with the higher probability. The final captions are marked in \textbf{bold}. }
\label{fig:generated_example_mscoco}
\end{figure*}
Our work uses the LSTM implementation of \cite{donahue2015long} on Caffe framework. All of our experiments were conducted on Ubuntu 14.04, 16G RAM and single Titan X GPU with 12G memory. Our LSTMs use 1000 hidden units and weights initialized uniformly from [-0.08, 0.08].  The batch sizes are 150, 100, 100, 32 for Bi-LSTM, Bi-S-LSTM, Bi-F-LSTM and Bi-LSTM (VGG) models respectively. Models were trained with learning rate $\eta=0.01$ (except $\eta=0.005$ for Bi-LSTM (VGG)), weight decay $\lambda$ is 0.0005 and we used momentum 0.9. Each model is trained for 18$\sim$35 epochs with early stopping. The code for this work can be found at \emph{\url{ https://github.com/deepsemantic/image_captioning}}.

\subsection{Evaluation Metrics}

We evaluate our models on two tasks: caption generation and image-sentence retrieval. In caption generation, we follow previous work to use BLEU-N (N=1,2,3,4) scores \cite{papineni2002bleu}:
\begin{equation}
B_N=\min(1,e^{1-\frac{r}{c}})\cdot e^{\frac{1}{N}\sum_{n=1}^N\log p_n}
\end{equation}
where $r$, $c$ are the length of reference sentence and generated sentence, $p_n$ is the modified \emph{$n$-gram} precisions.
We also report METETOR \cite{lavie2014meteor} and CIDEr \cite{vedantam2015cider} scores for further comparison. In image-sentence retrieval (image query sentence and vice versa), we adopt R@K (K=1,5,10) and Med $r$ as evaluation metrics. R@K is the recall rate R at top K candidates and Med $r$ is the median rank of the first retrieved ground-truth image and sentence. All mentioned metric scores are computed by MSCOCO caption evaluation server\footnote{https://github.com/tylin/coco-caption}, which is commonly used for image captioning challenge\footnote{http://mscoco.org/home/}.

\subsection{Visualization and Qualitative Analysis}
\label{sec.vis}
The aim of this set experiment is to visualize the properties of proposed bidirectional LSTM model and explain how it works in generating sentence word by word over time.

First, we examine the temporal evolution of internal gate states and understand how bidirectional LSTM units retain valuable context information and attenuate unimportant information. Figure \ref{fig:gates} shows input and output data, the pattern of three sigmoid gates (input, forget and output) as well
as cell states. We can clearly see that dynamic states are periodically distilled to units from time step $t=0$ to $t=11$. At $t=0$, the input data are sigmoid modulated to input gate $\mathbf{i}(t)$ where values lie within in [0,1]. At this step, the values of forget gates $\mathbf{f}(t)$ of different LSTM units are zeros. Along with the increasing of time step, forget gate starts to decide which unimportant information should be forgotten, meanwhile, to retain those useful information. Then the memory cell states $\mathbf{c}(t)$ and output gate $\mathbf{o}(t)$ gradually absorb the valuable context information over time and make a rich representation $\mathbf{h}(t)$ of the output data.

Next, we examine how visual and textual features are embedded to common semantic space and used to predict word over time. Figure \ref{fig:T-M-LSTM} shows the evolution of hidden units at different layers. For T-LSTM layer, units are conditioned by textual context from the past and future. It performs as the encoder of forward and backward sentences. At M-LSTM layer, LSTM units are conditioned by both visual and textual context. It learns the correlations between input word sequence and visual information that encoded by CNN. At given time step, by removing unimportant information that make less contribution to correlate input word and visual context, the units tend to appear sparsity pattern and learn more discriminative representations from inputs. At higher layer, embedded multimodal representations are used to compute the probability distribution of next predict word with Softmax. It should be noted, for given image, the number of words in generated sentence from forward and backward direction can be different.

Figure \ref{fig:generated_example_mscoco} presents some example images with generated captions. From it we found some interesting patterns of bidirectional captions: 
(1) \textbf{Cover different semantics}, for example, in (b) forward sentence captures ``\emph{couch}'' and ``\emph{table}'' while backward one describes ``\emph{chairs}'' and ``\emph{table}''. 
(2) \textbf{Describe static scenario and infer dynamics}, in (a) and (d), one caption describes the static scene, and the other one presents the potential action or motion that possibly happen in the next time step. 
(3) \textbf{Generate novel sentences}, from generated captions, we found that a significant proportion (88\% by randomly select 1000 images on MSCOCO validation set) of generated sentences are novel (not appear in training set). But generated sentences are highly similar to ground-truth captions, for example in (d), forward caption is similar to one of ground-truth captions (\emph{``A passenger train that is pulling into a station''}) and backward caption is similar to ground-truth caption (\emph{``a train is in a tunnel by a station''}). It illustrates that our model has strong capability in learning visual-language correlation and generating novel sentences.

\begin{table*}[t]
\centering
\small
\caption{Performance comparison on BLEU-N(high score is good). The superscript ``A'' means the visual model is AlexNet (or similar network), ``V'' is VggNet, ``G'' is GoogleNet, ``-'' indicates unknown value,  ``$\ddagger$'' means different data splits\protect\footnotemark. The best results are marked in $\mathbf{bold}$ and the second best results with \underline{underline}. The superscripts are also applicable to Table \ref{tab:recall}.}
\begin{tabular}{|l||p{0.7cm}p{0.7cm}p{0.7cm}p{0.7cm}||p{0.7cm}p{0.7cm}p{0.7cm}p{0.7cm}||p{0.7cm}p{0.7cm}p{0.7cm}p{0.7cm}|}
\hline
\multicolumn{1}{|l||}{} & \multicolumn{4}{c||}{\textbf{Flickr8K}}    & \multicolumn{4}{c||}{\textbf{Flickr30K}}   & \multicolumn{4}{c|}{\textbf{MSCOCO}} \\ \hline
\multicolumn{1}{|l||}{Models}   & \multicolumn{1}{l}{\textbf{B-1}} & \multicolumn{1}{l}{\textbf{B-2}} & \multicolumn{1}{l}{\textbf{B-3}} & \multicolumn{1}{l||}{\textbf{B-4}} & \multicolumn{1}{l}{\textbf{B-1}} & \multicolumn{1}{l}{\textbf{B-2}} & \multicolumn{1}{l}{\textbf{B-3}} & \multicolumn{1}{l||}{\textbf{B-4}}  & \multicolumn{1}{l}{\textbf{B-1}}  & \multicolumn{1}{l}{\textbf{B-2}}  & \multicolumn{1}{l}{\textbf{B-3}}  & \multicolumn{1}{l|}{\textbf{B-4}}  \\ \hline
NIC\cite{vinyals2015show}$^{G,\ddagger}$   & 63 & 41 & 27.2 & -  & $\underline{66.3}$    & 42.3    & 27.7    & 18.3    & 66.6    & 46.1    & 32.9    & 24.6    \\
LRCN\cite{donahue2015long}$^{A,\ddagger}$  & -  & -  & -  & -  & 58.8    & 39.1    & 25.1    & 16.5    & 62.8    & 44.2    & 30.4    & -  \\
DeepVS\cite{karpathy2015deep}$^{V}$& 57.9    & 38.3    & 24.5    & 16 & 57.3    & 36.9    & 24.0    & 15.7    & 62.5    & 45 & 32.1    & 23 \\
m-RNN\cite{mao2014deep}$^{A,\ddagger}$ & 56.5 & 38.6  & 25.6 & 17.0  & 54 & 36 & 23 & 15 & - & - & - & - \\
m-RNN\cite{mao2014deep}$^{V,\ddagger}$ & -  & -  & -  & -  & 60 & 41 & 28 & 19 & 67 & 49 & 35 & $\underline{25}$ \\
Hard-Attention\cite{xu2015show}$^{V}$ &  $\mathbf{67}$ & $\underline{45.7}$  & $\underline{31.4}$ & $\underline{21.3}$ & $\mathbf{66.9}$ &  $\mathbf{43.9}$ &  $\mathbf{29.6}$ &  $\mathbf{19.9}$ &  $\mathbf{71.8}$ &  $\mathbf{50.4}$ & $\mathbf{35.7}$ & $\mathbf{25}$ \\ \hline
Bi-LSTM$^{A}$&  61.9  & 43.3   &  29.7  &  20.0  & 58.9   & 39.3   & 25.9   & 17.1   & 63.4   &  44.7  &  30.6  & 20.6   \\
Bi-S-LSTM$^{A}$&  64.2  & 44.3   &  29.2  &  18.6  & 59.5   & 40.3   & 26.9   & 17.9   &  63.7   &  45.7  &  31.8  & 21.9\\
Bi-F-LSTM$^{A}$&  63.0  & 43.7   &  29.2  &  19.1  & 58.6    & 39.2   & 26.0   & 17.4  & 63.5   &  44.8  &  30.7  & 20.6  \\
Bi-LSTM$^{V}$&  $\underline{65.5}$  &  $\mathbf{46.8}$   &   $\mathbf{32.0}$  &   $\mathbf{21.5}$  & 62.1    & $\underline{42.6}$   & $\underline{28.1}$   & $\underline{19.3}$  & $\underline{67.2}$   &  $\underline{49.2}$  &  $\underline{35.2}$  & 24.4  \\ \hline
\end{tabular}
\label{tab:BLEU}
\vspace{-2ex}
\end{table*}

%%%%  Results of Caption Generation %%%%
\subsection{Results on Caption Generation}
\label{sec.com_generation}
Now, we compare with state-of-the-art methods. Table \ref{tab:BLEU} presents the comparison results in terms of BLEU-N. Our approach achieves very competitive performance on evaluated datasets although with less powerful AlexNet visual model. We can see that increase the depth of LSTM is beneficial on generation task. Deeper variant models mostly obtain better performance compare to Bi-LSTM, but they are inferior to latter one in B-3 and B-4 on Flickr8K. We conjecture it should be the reason that Flick8K is a relatively small dataset which suffers difficulty in training deep models with limited data. One of interesting facts we found is that by stacking multiple LSTM layers is generally superior to LSTM with fully connected transition layer although Bi-S-LSTM needs more training time. By replacing AlexNet with VggNet brings significant improvements on all BLEU evaluation metrics. We should be aware of that a recent interesting work \cite{xu2015show} achieves the best results by integrating attention mechanism \cite{lecun2015deep,xu2015show} on this task. Although we believe incorporating such powerful mechanism into our framework can make further improvements, note that our current model Bi-LSTM$^V$ achieves the best or second best results on most of metrics while the small gap in performance between our model and Hard-Attention$^V$ \cite{xu2015show} is existed. 

\begin{figure}[h]
\centering
  \subfigure[METEOR score]{
    \includegraphics[width=0.225\textwidth]{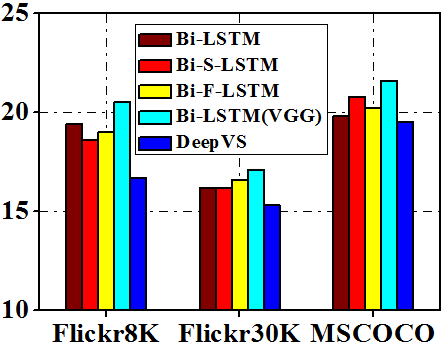}}
  \subfigure[CIDEr score]{
    \includegraphics[width=0.225\textwidth]{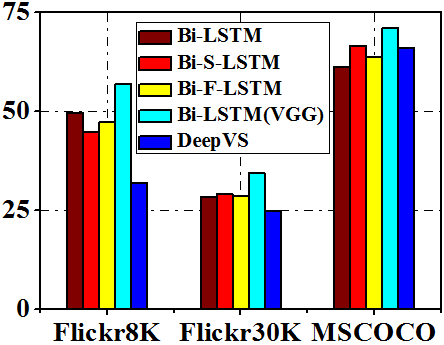}}
\vspace{-2ex}
\caption{METEOR/CIDEr scores on different datasets.}
\label{fig:METEOR_CIDEr}
\end{figure}

\footnotetext{On MSCOCO dataset, NIC uses 4K images for validation and test. LRCN randomly selects 5K images from MSCOCO validation set for validation and test. m-RNN uses 4K images for validation and 1K as test.}
The further comparison on METEOR and CIDEr scores is plotted in Figure \ref{fig:METEOR_CIDEr}. Without integrating object detection and more powerful vision model, our model (Bi-LSTM$^A$) outperforms DeepVS$^V$\cite{karpathy2015deep} in a certain margin. It achieves 19.4/49.6 on Flickr 8K (compare to 16.7/31.8 of DeepVS$^V$) and 16.2/28.2 on Flickr30K (15.3/24.7 of DeepVS$^V$). On MSCOCO, our Bi-S-LSTM$^A$ obtains 20.8/66.6 for METEOR/CIDEr, which exceeds 19.5/66.0 in DeepVS$^V$. 
\begin{table*}[htb]
\centering
\small
\caption{Comparison with state-of-the-art methods on R@K (high is good) and Med r (low is good). All scores are computed by averaging the results of forward and backward results. ``$+O$'' means the approach with additional object detection.  }
\label{tab:recall}
\begin{tabular}{|c|c|cp{0.9cm}p{0.9cm}p{0.9cm}|p{0.9cm}p{0.9cm}p{0.9cm}p{0.9cm}|}
\hline
 &  & \multicolumn{4}{c|}{Image to Sentence} & \multicolumn{4}{c|}{Sentence to Image} \\ \cline{3-10} 
 Datasets& Methods & \multicolumn{1}{l}{R@1} & R@5 & R@10 & \multicolumn{1}{l|}{Med r} & R@1 & R@5 & R@10 & \multicolumn{1}{l|}{Med r} \\ \hline\hline
\multirow{11}{*}{Flickr 8K} 
 & DeViSE\cite{frome2013devise} & 4.8 & 16.5 & 27.3 & 28 & 5.9 & 20.1 & 29.6 & 29 \\
 & SDT-RNN\cite{socher2014grounded} & 4.5 & 18.0 & 28.6 & 32 & 6.1 & 18.5 & 29.0 & 29 \\
 & DeFrag\cite{karpathy2014deep}$^{+O}$ & 12.6 & 32.9 & 44.0 & 14 & 9.7 & 29.6 & 42.5 & 15 \\
 & Kiros et al. \cite{kiros2014unifying}$^{A}$ & 13.5 & 36.2 & 45.7 & 13 & 10.4 & 31.0 & 43.7 & 14 \\
 & Kiros et al. \cite{kiros2014unifying}$^{V}$ & 18 & 40.9 & 55 & 8 & 12.5 & 37 & 51.5 & 10 \\
 & m-RNN\cite{mao2014deep}$^{A}$ & 14.5 & 37.2 & 48.5 & 11 & 11.5 & 31.0 & 42.4 & 15 \\
 & Mind's Eye\cite{chen2015mind}$^{V}$ & 17.3 & 42.5 & 57.4 & 7 & 15.4 & 40.6 & 50.1 & 8 \\
 & DeepVS\cite{karpathy2015deep}$^{+O,V}$ & 16.5 & 40.6 & 54.2 & 7.6 & 11.8 & 32.1 & 44.7 & 12.4 \\
 & NIC\cite{vinyals2015show}$^{G}$ & 20 & - & 60 & 6 & 19 & - & $\mathbf{64}$ & 5 \\ \cline{2-10}
 & Bi-LSTM$^{A}$ & 21.3 & 44.7 & 56.5 & 6.5 & 15.1 & 37.8 & 50.1 & 9 \\
 & Bi-S-LSTM$^{A}$ & 19.6 & 43.7 & 55.7 & 7 & 14.5 & 36.4 & 48.3 & 10.5 \\
 & Bi-F-LSTM$^{A}$ & 19.9 & 44.0 & 56.0 & 7 & 14.9 & 37.4 & 49.8 & 10 \\
 & Bi-LSTM$^{V}$ & $\mathbf{29.3}$ & $\mathbf{58.2}$ & $\mathbf{69.6}$ & $\mathbf{3}$ & $\mathbf{19.7}$ & $\mathbf{47.0}$ & 60.6 & $\mathbf{5}$ \\ \hline \hline
\multirow{11}{*}{Flickr 30K} 
 & DeViSE\cite{frome2013devise} & 4.5 & 18.1 & 29.2 & 26 & 6.7 & 21.9 & 32.7 & 25 \\
 & SDT-RNN\cite{socher2014grounded} & 9.6 & 29.8 & 41.1 & 16 & 8.9 & 29.8 & 41.1 & 16 \\
 & Kiros et al. \cite{kiros2014unifying}$^{A}$ & 14.8 & 39.2 & 50.9 & 10 & 11.8 & 34.0 & 46.3 & 13 \\ 
 & Kiros et al. \cite{kiros2014unifying}$^{V}$ & 23.0 & 50.7 & 62.9 & 5 & 16.8 & 42.0 & 56.5 & 8 \\
 & LRCN\cite{donahue2015long}$^{A}$ & 14 & 34.9 & 47 & 11 & - & - & - & - \\
 & NIC\cite{vinyals2015show}$^{G}$ & 17 & - & 56 & 7 & 17 & - &57 & 8 \\
 & m-RNN\cite{mao2014deep}$^{A}$ & 18.4 & 40.2 & 50.9 & 10 & 12.6 & 31.2 & 41.5 & 16 \\
 & Mind's Eye\cite{chen2015mind}$^{V}$ & 18.5 & 45.7 & 58.1 & 7 & 16.6 & 42.5 &  $\mathbf{58.9}$ & 8 \\
 & DeFrag \cite{karpathy2014deep}$^{+O}$ & 16.4 & 40.2 & 54.7 & 8 & 10.3 & 31.4 & 44.5 & 13 \\
 & DeepVS\cite{karpathy2015deep}$^{+O,V}$ & 22.2 & 48.2 & 61.4 & $4.8$ & 15.2 & 37.7 & 50.5 & 9.2 \\ \cline{2-10}
 & Bi-LSTM$^{A}$ & 18.7 & 41.2 & 52.6 & 8 & 14.0 & 34.0 & 44.0 & 14 \\
 & Bi-S-LSTM$^{A}$ & 21 & 43.0 & 54.1 & 7 & 15.1 & 35.3 & 46.0 & 12 \\
 & Bi-F-LSTM$^{A}$ & 20 & 44.4 & 55.2 & 7 & 15.1 & 35.8 & 46.8 & 12 \\
 & Bi-LSTM$^{V}$ & $\mathbf{28.1}$ & $\mathbf{53.1}$ & $\mathbf{64.2}$ & $\mathbf{4}$ & $\mathbf{19.6}$ & $\mathbf{43.8}$ & 55.8 & $\mathbf{7}$ \\ \hline \hline
\multirow{5}{*}{MSCOCO} & DeepVS\cite{karpathy2015deep}$^{+O,V}$ & \multicolumn{1}{l}{16.5} & 39.2 & 52.0 & \multicolumn{1}{l|}{9} & 10.7 & 29.6 & 42.2 & 14.0 \\ \cline{2-10} 
 & Bi-LSTM$^{A}$ & 10.8 & 28.1 & 38.9 & 18 & 7.8 & 22.4 & 32.8 & 24 \\
 & Bi-S-LSTM$^{A}$ & 13.4 & 33.1 & 44.7 & 13 & 9.4 & 26.5 & 37.7 & 19 \\
 & Bi-F-LSTM$^{A}$ & 11.2 & 30 & 41.2 & 16 & 8.3 & 24.9 & 35.1 & 22 \\
 & Bi-LSTM$^{V}$ & $\mathbf{16.6}$ & $\mathbf{39.4}$ & $\mathbf{52.4} $ & $\mathbf{9}$ & $\mathbf{11.6}$ & $\mathbf{30.9}$ & $\mathbf{43.4}$ & $\mathbf{13}$ \\ \hline
\end{tabular}
\vspace{-2ex}
\end{table*}
\vspace{-3ex}
\subsection{Results on Image-Sentence Retrieval}
\label{sec.retrieval}
\begin{figure*}[htb]
\begin{table}[H]
\begin{tabular}{ccccc}
\fcolorbox{red}{white!20}{
\begin{minipage}{0.15\textwidth}
snowboarder \\ jumping
\end{minipage} }
&
\begin{minipage}{0.18\textwidth}
\centering
\fcolorbox{green}{white!20}{\includegraphics[width=3cm,height=2cm]{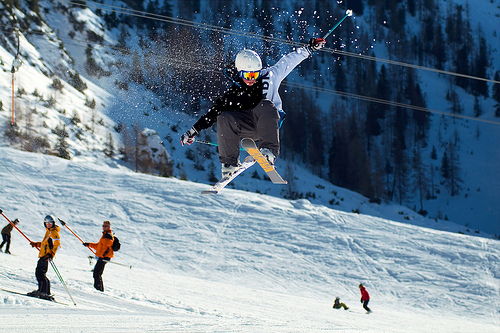}}
\end{minipage} 
&
\begin{minipage}{0.18\textwidth}
\centering
\fcolorbox{green}{white!20}{\includegraphics[width=3cm,height=2cm]{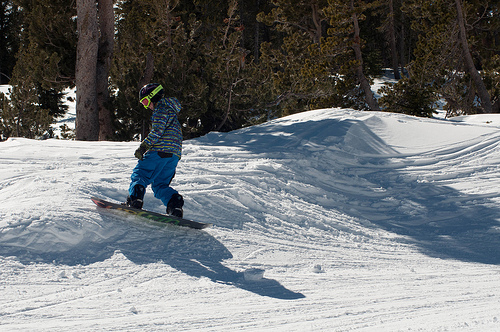}}
\end{minipage} 
&
\begin{minipage}{0.18\textwidth}
\centering
\fcolorbox{green}{white!20}{\includegraphics[width=3cm,height=2cm]{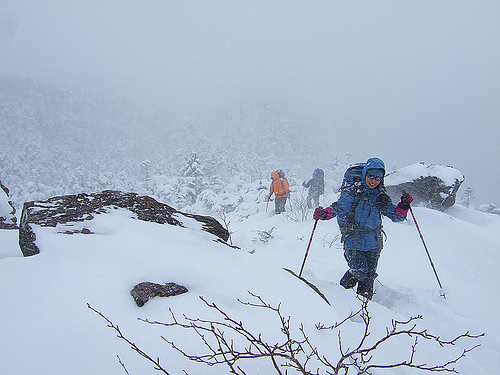}}
\end{minipage} 
&
\begin{minipage}{0.18\textwidth}
\centering
\fcolorbox{green}{white!20}{\includegraphics[width=3cm,height=2cm]{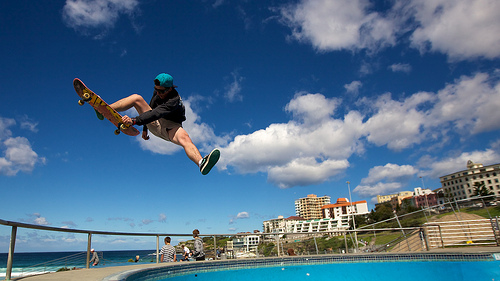}}
\end{minipage} \vspace{0.5ex}\\ \vspace{0.5ex}

\fcolorbox{red}{white!20}{
\begin{minipage}{0.15\textwidth}
A man practices \\ boxing
\end{minipage} }
&
\begin{minipage}{0.18\textwidth}
\centering
\fcolorbox{green}{white!20}{\includegraphics[width=3cm,height=2cm]{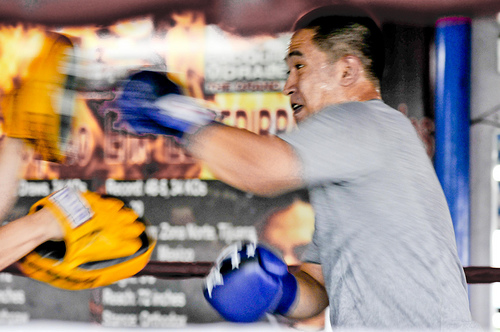}}
\end{minipage} 
&
\begin{minipage}{0.18\textwidth}
\centering
\fcolorbox{green}{white!20}{\includegraphics[width=3cm,height=2cm]{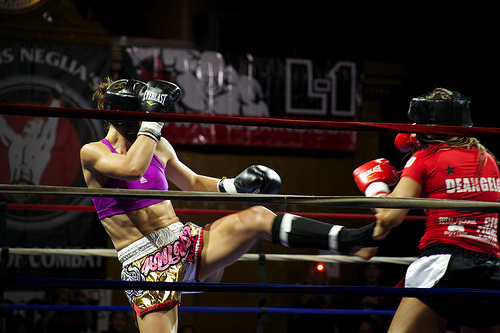}}
\end{minipage} 
&
\begin{minipage}{0.18\textwidth}
\centering
\fcolorbox{green}{white!20}{\includegraphics[width=3cm,height=2cm]{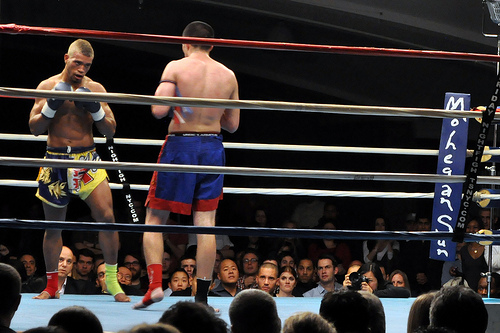}}
\end{minipage} 
&
\begin{minipage}{0.18\textwidth}
\centering
\fcolorbox{green}{white!20}{\includegraphics[width=3cm,height=2cm]{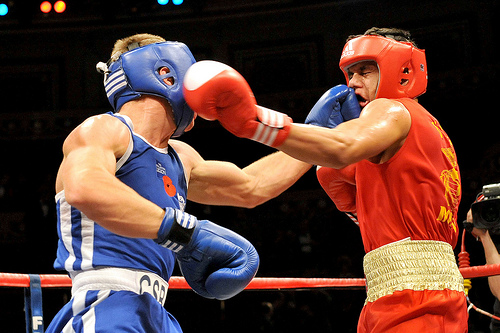}}
\end{minipage}  \vspace{0.5ex} \\ \Xhline{1.1pt} \vspace{-1.5ex} \\  \vspace{0.125ex}
\begin{minipage}{0.18\textwidth}
\centering
\fcolorbox{red}{white!20}{\includegraphics[width=3cm,height=2cm]{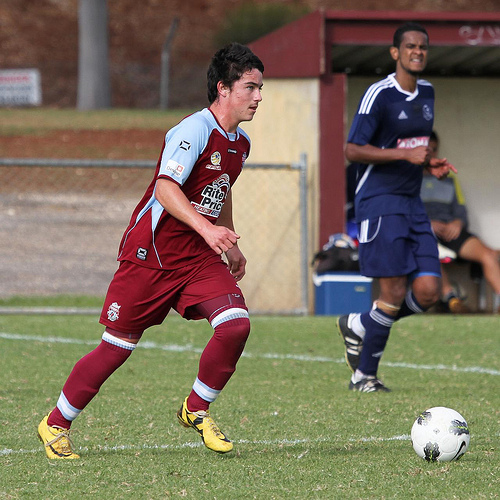}}
\end{minipage} 
&
\fcolorbox{green}{white!20}{
\begin{minipage}{0.15\textwidth}
Two guys one in a red uniform and one in a blue uniform playing soccer
\end{minipage} }
&
\fcolorbox{green}{white!20}{
\begin{minipage}{0.15\textwidth}
A soccer player tackles a player from the other team
\end{minipage} }
&
\fcolorbox{green}{white!20}{
\begin{minipage}{0.15\textwidth}
Two young men tackle an opponent during a scrimmage football game
\end{minipage} }
&
\fcolorbox{green}{white!20}{
\begin{minipage}{0.15\textwidth}
A football player preparing a football for a field goal kick, while his teammates can coach watch him
\end{minipage}} \\ \vspace{0.25ex}

\begin{minipage}{0.18\textwidth}
%\centering
\fcolorbox{red}{white!20}{\includegraphics[width=3cm,height=2cm]{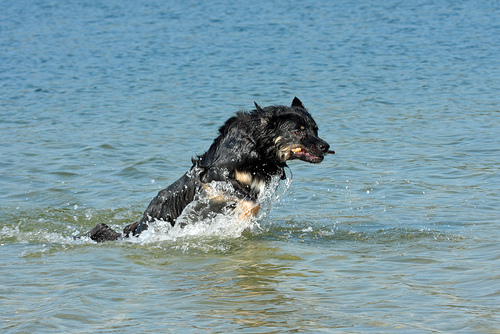}}
\end{minipage} 
&
\fcolorbox{green}{white!20}{
\begin{minipage}{0.15\textwidth}
Black dog jumping out of the water with a stick in his mouth
\end{minipage}}
&
\fcolorbox{green}{white!20}{
\begin{minipage}{0.15\textwidth}
A black dog jumps in a body of water with a stick in his mouth
\end{minipage}}
&
\fcolorbox{green}{white!20}{
\begin{minipage}{0.15\textwidth}
A black dog is swimming through water
\end{minipage}}
&
\fcolorbox{green}{white!20}{
\begin{minipage}{0.15\textwidth}
A black dog swimming in the water with a tennis ball in his mouth
\end{minipage}}

\end{tabular}
\end{table} 
\vspace{-2ex}
\caption{Examples of image retrieval (top) and caption retrieval (bottom) with Bi-S-LSTM on Flickr30K validation set. Queries are marked with red color and top-4 retrieved results are marked with green color.}
\label{fig:retr_example}
\end{figure*}

For retrieval evaluation, we focus on image to sentence retrieval and vice versa. This is an instance of cross-modal retrieval \cite{feng2014cross,costa2014role,jiang2015deep} which has been a hot research subject in multimedia field.  Table \ref{tab:recall} illustrates our results on different datasets. The performance of our models exceeds those compared methods on most of metrics or matching existing results. In a few metrics, our model didn't show better result than Mind's Eye \cite{chen2015mind} which combined image and text features in ranking (it makes this task more like multimodal retrieval) and NIC \cite{vinyals2015show} which employed more powerful vision model, large beam size and model ensemble. While adopting more powerful visual model VggNet results in significant improvements across all metrics, with less powerful AlexNet model, our results are still competitive on some metrics, e.g. R@1, R@5 on Flickr8K and Flickr30K. We also note that on relatively small dataset Filckr8K, shallow model performs slightly better than deeper ones on retrieval task, which in contrast with the results on the other two datasets. As we explained before, we think deeper LSTM architectures are better suited for ranking task on large datasets which provides enough training data for more complicate model training, otherwise, overfitting occurs. By increasing data variations with our implemented data augmentation techniques can alleviate it in a certain degree. But we foresee further significant improvement gains as training example grows, by reducing reliance on augmentation with fresh data. Figure \ref{fig:retr_example} presents some examples of retrieval experiments. For each caption (image) query, sensible images and descriptive captions are retrieved. It shows our models captured the visual-textual correlation for image and sentence ranking.% are learned by . %Although some retrieved results are not the corresponding ground-truth image or captions. For instance, in third query, only the first cation is one of the five truth captions, but the other two results are very positive and similar to its truth caption(i.e \emph{``Two men on opposing teams race toward a soccer ball''}).

\subsection{Discussion}

\textbf{Efficiency}. In addition to showing superior performance, our models also possess high computational efficiency. Table \ref{tab:time_costs} presents the computational costs of proposed models. We randomly select 10 images from Flickr8K validation set, and perform caption generation and image to sentence retrieval test for 5 times respectively. The table shows the averaged time costs across 5 test results. The time cost of network initialization is excluded. The costs of caption generation includes: computing image feature, sampling bidirectional captions, computing the final caption. The time costs for retrieval considers: computing image-sentence pair scores (totally 10 $\times$ 50 pairs), ranking sentences for each image query. As can be seen from Table \ref{tab:BLEU}, \ref{tab:recall} and \ref{tab:time_costs}, deep models have only slightly higher time consumption but yield significant improvements and our proposed Bi-F-LSTM can strike the balance between performance and efficiency.
\begin{table}
\centering
\caption{Time costs for testing 10 images on Flickr8K}
\label{tab:time_costs}
\begin{tabular}{|c|c|c|c|}
\hline
       & Bi-LSTM & Bi-S-LSTM & Bi-F-LSTM \\ \hline
Generation      & 0.93s   & 1.1s      & 0.97s     \\ \hline
Retrieval       & 5.62s   & 7.46s     & 5.69s     \\ \hline
\end{tabular}
\vspace{-3ex}
\end{table}

\textbf{Challenges in exact comparison}. It is challenging to make a direct, extract comparison with related methods due to the differences in dataset division on MSCOCO. In principle, testing on smaller validation set can lead to better results, particularly in retrieval task. Since we strictly follow dataset splits as in \cite{karpathy2015deep}, we compare to it in most cases. Another challenge is the visual model that utilized for encoding image inputs. Different models are employed in different works, 
%as more powerful model (i.e GoogleNet, VggNet) achieves better performance. 
to make a fair and comprehensive comparison, we select commonly used AlexNet and VggNet in our work.

\section{Conclusions}
We proposed a bidirectional LSTM model that generates descriptive sentence for image by taking both history and future context into account. We further designed deep bidirectional LSTM architectures to embed image and sentence at high semantic space for learning visual-language models.  We also qualitatively visualized internal states of proposed model to understand how multimodal LSTM generates word at consecutive time steps. The effectiveness, generality and robustness of proposed models were evaluated on numerous datasets. Our models achieve highly completive or state-of-the-art results on both generation and retrieval tasks. Our future work will focus on exploring more sophisticated language representation (e.g. word2vec) and incorporating multitask learning and attention mechanism into our model. We also plan to apply our model to other sequence learning tasks such as text recognition and video captioning.
\bibliographystyle{plain}
\bibliography{references}

\begin{thebibliography}{10}

\bibitem{bahdanau2014neural}
D.~Bahdanau, K.~Cho, and Y.~Bengio.
\newblock Neural machine translation by jointly learning to align and
  translate.
\newblock {\em ICLR}, 2015.

\bibitem{chen2015mind}
X.~Chen and C.~Lawrence Zitnick.
\newblock Mind's eye: A recurrent visual representation for image caption
  generation.
\newblock In {\em CVPR}, pages 2422--2431, 2015.

\bibitem{cho2014learning}
K.~Cho, B.~Van Merri{\"e}nboer, C.~Gulcehre, D.~Bahdanau, F.~Bougares,
  H.~Schwenk, and Y.~Bengio.
\newblock Learning phrase representations using rnn encoder-decoder for
  statistical machine translation.
\newblock {\em EMNLP}, 2014.

\bibitem{donahue2015long}
J.~Donahue, L.~A. Hendricks, S.~Guadarrama, M.~Rohrbach, S.~Venugopalan,
  K.~Saenko, and T.~Darrell.
\newblock Long-term recurrent convolutional networks for visual recognition and
  description.
\newblock In {\em CVPR}, pages 2625--2634, 2015.

\bibitem{fang2015captions}
H.~Fang, S.~Gupta, F.~Iandola, R.~Srivastava, L.~Deng, P.~Doll{\'a}r, J.~Gao,
  X.~He, M.~Mitchell, and J.~Platt.
\newblock From captions to visual concepts and back.
\newblock In {\em CVPR}, pages 1473--1482, 2015.

\bibitem{feng2014cross}
F.~Feng, X.~Wang, and R.~Li.
\newblock Cross-modal retrieval with correspondence autoencoder.
\newblock In {\em ACMMM}, pages 7--16. ACM, 2014.

\bibitem{frome2013devise}
A.~Frome, G.~Corrado, J.~Shlens, S.~Bengio, J.~Dean, and T.~Mikolov.
\newblock Devise: A deep visual-semantic embedding model.
\newblock In {\em NIPS}, pages 2121--2129, 2013.

\bibitem{graves2013speech}
A.~Graves, A.~Mohamed, and G.~E. Hinton.
\newblock Speech recognition with deep recurrent neural networks.
\newblock In {\em ICASSP}, pages 6645--6649. IEEE, 2013.

\bibitem{jia2014caffe}
Y.~Jia, E.~Shelhamer, J.~Donahue, S.~Karayev, J.~Long, R.~Girshick,
  S.~Guadarrama, and T.~Darrell.
\newblock Caffe: Convolutional architecture for fast feature embedding.
\newblock In {\em ACMMM}, pages 675--678. ACM, 2014.

\bibitem{karpathy2014deep}
A.~Karpathy, A.~Joulin, and F-F. Li.
\newblock Deep fragment embeddings for bidirectional image sentence mapping.
\newblock In {\em NIPS}, pages 1889--1897, 2014.

\bibitem{karpathy2015deep}
A.~Karpathy and F-F. Li.
\newblock Deep visual-semantic alignments for generating image descriptions.
\newblock In {\em CVPR}, pages 3128--3137, 2015.

\bibitem{kiros2014multimodal}
R.~Kiros, R.~Salakhutdinov, and R.~Zemel.
\newblock Multimodal neural language models.
\newblock In {\em ICML}, pages 595--603, 2014.

\bibitem{kiros2014unifying}
R.~Kiros, R.~Salakhutdinov, and R.~Zemel.
\newblock Unifying visual-semantic embeddings with multimodal neural language
  models.
\newblock {\em arXiv preprint arXiv:1411.2539}, 2014.

\bibitem{krizhevsky2012imagenet}
A.~Krizhevsky, I.~Sutskever, and G.~E. Hinton.
\newblock Imagenet classification with deep convolutional neural networks.
\newblock In {\em NIPS}, pages 1097--1105, 2012.

\bibitem{kulkarni2013babytalk}
G.~Kulkarni, V.~Premraj, V.~Ordonez, S.~Dhar, S.~Li, Y.~Choi, A.~C. Berg, and
  T.~Berg.
\newblock Babytalk: Understanding and generating simple image descriptions.
\newblock {\em IEEE Trans. on Pattern Analysis and Machine Intelligence(PAMI)},
  35(12):2891--2903, 2013.

\bibitem{kuznetsova2012collective}
P.~Kuznetsova, V.~Ordonez, A.~C. Berg, T.~Berg, and Y.~Choi.
\newblock Collective generation of natural image descriptions.
\newblock In {\em ACL}, volume~1, pages 359--368. ACL, 2012.

\bibitem{kuznetsova2014treetalk}
P.~Kuznetsova, V.~Ordonez, T.~Berg, and Y.~Choi.
\newblock Treetalk: Composition and compression of trees for image
  descriptions.
\newblock {\em Trans. of the Association for Computational Linguistics(TACL)},
  2(10):351--362, 2014.

\bibitem{lavie2014meteor}
M.~Lavie.
\newblock Meteor universal: language specific translation evaluation for any
  target language.
\newblock {\em ACL}, page 376, 2014.

\bibitem{lecun2015deep}
Y.~LeCun, Y.~Bengio, and G.~E. Hinton.
\newblock Deep learning.
\newblock {\em Nature}, 521(7553):436--444, 2015.

\bibitem{li2011composing}
S.~Li, G.~Kulkarni, T.~L. Berg, A.~C. Berg, and Y.~Choi.
\newblock Composing simple image descriptions using web-scale n-grams.
\newblock In {\em CoNLL}, pages 220--228. ACL, 2011.

\bibitem{lin2014microsoft}
T-Y. Lin, M.~Maire, S.~Belongie, J.~Hays, P.~Perona, D.~Ramanan, P.~Doll{\'a}r,
  and C.~Zitnick.
\newblock Microsoft coco: Common objects in context.
\newblock In {\em ECCV}, pages 740--755. Springer, 2014.

\bibitem{lu2014rapid}
Xin Lu, Zhe Lin, Hailin Jin, Jianchao Yang, and James~Z Wang.
\newblock Rapid: Rating pictorial aesthetics using deep learning.
\newblock In {\em ACMMM}, pages 457--466. ACM, 2014.

\bibitem{mao2014deep}
J.~H. Mao, W.~Xu, Y.~Yang, J.~Wang, Z.~H. Huang, and A.~Yuille.
\newblock Deep captioning with multimodal recurrent neural networks (m-rnn).
\newblock {\em ICLR}, 2015.

\bibitem{mikolov2010recurrent}
T.~Mikolov, M.~Karafi{\'a}t, L.~Burget, J.~Cernock{\`y}, and S.~Khudanpur.
\newblock Recurrent neural network based language model.
\newblock In {\em INTERSPEECH}, volume~2, page~3, 2010.

\bibitem{mikolov2011extensions}
T.~Mikolov, S.~Kombrink, L.~Burget, J.~H. {\v{C}}ernock{\`y}, and S.~Khudanpur.
\newblock Extensions of recurrent neural network language model.
\newblock In {\em ICASSP}, pages 5528--5531. IEEE, 2011.

\bibitem{mitchell2012midge}
M.~Mitchell, X.~Han, J.~Dodge, A.~Mensch, A.~Goyal, A.~Berg, K.~Yamaguchi,
  T.~Berg, K.~Stratos, and H.~Daum{\'e} III.
\newblock Midge: Generating image descriptions from computer vision detections.
\newblock In {\em ACL}, pages 747--756. ACL, 2012.

\bibitem{ngiam2011multimodal}
J.~Ngiam, A.~Khosla, M.~Kim, J.~Nam, H.~Lee, and A.~Ng.
\newblock Multimodal deep learning.
\newblock In {\em ICML}, pages 689--696, 2011.

\bibitem{papineni2002bleu}
K.~Papineni, S.~Roukos, T.~Ward, and W.~Zhu.
\newblock Bleu: a method for automatic evaluation of machine translation.
\newblock In {\em ACL}, pages 311--318. ACL, 2002.

\bibitem{pascanu2013construct}
R.~Pascanu, C.~Gulcehre, K.~Cho, and Y.~Bengio.
\newblock How to construct deep recurrent neural networks.
\newblock {\em arXiv preprint arXiv:1312.6026}, 2013.

\bibitem{costa2014role}
J.~C. Pereira, E.~Coviello, G.~Doyle, N.~Rasiwasia, G.~Lanckriet, R.~Levy, and
  N.~Vasconcelos.
\newblock On the role of correlation and abstraction in cross-modal multimedia
  retrieval.
\newblock {\em IEEE Trans. on Pattern Analysis and Machine Intelligence(PAMI)},
  36(3):521--535, 2014.

\bibitem{rashtchian2010collecting}
C.~Rashtchian, P.~Young, M.~Hodosh, and J.~Hockenmaier.
\newblock Collecting image annotations using amazon's mechanical turk.
\newblock In {\em NAACL HLT Workshop}, pages 139--147. Association for
  Computational Linguistics, 2010.

\bibitem{simonyan2014two}
K.~Simonyan and A.~Zisserman.
\newblock Two-stream convolutional networks for action recognition in videos.
\newblock In {\em NIPS}, pages 568--576, 2014.

\bibitem{simonyan2014very}
K.~Simonyan and A.~Zisserman.
\newblock Very deep convolutional networks for large-scale image recognition.
\newblock {\em arXiv preprint arXiv:1409.1556}, 2014.

\bibitem{socher2014grounded}
R.~Socher, A.~Karpathy, Q.~V. Le, C.~D. Manning, and A.~Y. Ng.
\newblock Grounded compositional semantics for finding and describing images
  with sentences.
\newblock {\em Trans. of the Association for Computational Linguistics(TACL)},
  2:207--218, 2014.

\bibitem{srivastava2012multimodal}
N.~Srivastava and R.~Salakhutdinov.
\newblock Multimodal learning with deep boltzmann machines.
\newblock In {\em NIPS}, pages 2222--2230, 2012.

\bibitem{sutskever2014sequence}
I.~Sutskever, O.~Vinyals, and Q.~V. Le.
\newblock Sequence to sequence learning with neural networks.
\newblock In {\em NIPS}, pages 3104--3112, 2014.

\bibitem{szegedy2015going}
C.~Szegedy, W.~Liu, Y.~Jia, P.~Sermanet, S.~Reed, D.~Anguelov, D.~Erhan,
  V.~Vanhoucke, and A.~Rabinovich.
\newblock Going deeper with convolutions.
\newblock In {\em CVPR}, pages 1--9, 2015.

\bibitem{vedantam2015cider}
R.~Vedantam, Z.~Lawrence, and D.~Parikh.
\newblock Cider: Consensus-based image description evaluation.
\newblock In {\em CVPR}, pages 4566--4575, 2015.

\bibitem{vinyals2015show}
O.~Vinyals, A.~Toshev, S.~Bengio, and D.~Erhan.
\newblock Show and tell: A neural image caption generator.
\newblock In {\em CVPR}, pages 3156--3164, 2015.

\bibitem{Wang:2015:DIY:2733373.2806219}
Zhangyang Wang, Jianchao Yang, Hailin Jin, Eli Shechtman, Aseem Agarwala,
  Jonathan Brandt, and Thomas~S. Huang.
\newblock Deepfont: Identify your font from an image.
\newblock In {\em ACMMM}, pages 451--459. ACM, 2015.

\bibitem{jiang2015deep}
X.Jiang, F.~Wu, X.~Li, Z.~Zhao, W.~Lu, S.~Tang, and Y.~Zhuang.
\newblock Deep compositional cross-modal learning to rank via local-global
  alignment.
\newblock In {\em ACMMM}, pages 69--78. ACM, 2015.

\bibitem{xu2015show}
K.~Xu, J.~Ba, R.~Kiros, A.~Courville, R.~Salakhutdinov, R.~Zemel, and
  Y.~Bengio.
\newblock Show, attend and tell: Neural image caption generation with visual
  attention.
\newblock {\em ICML}, 2015.

\bibitem{young2014image}
P.~Young, A.~Lai, M.~Hodosh, and J.~Hockenmaier.
\newblock From image descriptions to visual denotations: New similarity metrics
  for semantic inference over event descriptions.
\newblock {\em Trans. of the Association for Computational Linguistics(TACL)},
  2:67--78, 2014.

\bibitem{zaremba2014learning}
W.~Zaremba and I.~Sutskever.
\newblock Learning to execute.
\newblock {\em arXiv preprint arXiv:1410.4615}, 2014.

\bibitem{zeiler2014visualizing}
M.~D. Zeiler and R.~Fergus.
\newblock Visualizing and understanding convolutional networks.
\newblock In {\em ECCV}, pages 818--833. Springer, 2014.

\end{thebibliography}

%ACKNOWLEDGMENTS are optional
%\section{Acknowledgments}
\end{document}